\documentclass[10pt,journal,compsoc]{IEEEtran}

\ifCLASSOPTIONcompsoc
  \usepackage[nocompress]{cite}
\else
  \usepackage{cite}
\fi

\usepackage{xcolor}
\usepackage{graphicx}
\usepackage{amsmath}
\usepackage{acronym}
\usepackage{amssymb}
\usepackage{algorithmic}

\usepackage{subfigure}
\usepackage{subfloat}
\usepackage{multirow}
\usepackage{booktabs}

\usepackage{makecell}
\usepackage{array}
\usepackage{url}
\usepackage{commath}
\usepackage{textcomp}

\usepackage[pagebackref=true,breaklinks=true,colorlinks,bookmarks=false]{hyperref}
\usepackage{marvosym}

\usepackage[symbol]{footmisc}

\usepackage{xspace}
\usepackage{ragged2e}
\usepackage[labelfont=bf, font=footnotesize, justification=justified, format=plain]{caption}

\usepackage{xspace}
\usepackage{booktabs}
\usepackage{caption}[font={small}]

\makeatletter
\DeclareRobustCommand\onedot{\futurelet\@let@token\@onedot}
\def\@onedot{\ifx\@let@token.\else.\null\fi\xspace}

\def\eg{\emph{e.g}\onedot} 
\def\ie{\emph{i.e}\onedot}

\def\etal{\emph{et al}\onedot}
\makeatother

\newcommand{\name}{PointHPS\xspace}

\newcommand{\Fig}{Fig.\xspace}
\newcommand{\Tab}{Table\xspace}
\newcommand{\Sec}{Sec.\xspace}
\newcommand{\Supp}{Supplementary Material\xspace}

\begin{document}
\title{\name: Cascaded 3D Human Pose and Shape Estimation from Point Clouds}

\author{Zhongang Cai$^{\star, 1, 2}$ \quad
Liang Pan$^{\star, 1}$ \quad
Chen Wei$^{2}$ \quad
Wanqi Yin$^{2}$ \quad
Fangzhou Hong$^{1}$ \\
Mingyuan Zhang$^{1}$ \quad
Chen Change Loy$^{1}$ \quad
Lei Yang$^{2}$ \quad
Ziwei Liu$^{1, \href{mailto:ziwei.liu@ntu.edu.sg}{\textrm{\Letter}}}$%
\IEEEcompsocitemizethanks{
    \IEEEcompsocthanksitem $^{\star}$ indicates equal contributions.
    \IEEEcompsocthanksitem $^{1}$ S-Lab, Nanyang Technological University.
    \IEEEcompsocthanksitem $^{2}$ SenseTime Research.
    \IEEEcompsocthanksitem The corresponding author is Ziwei Liu: ziwei.liu@ntu.edu.sg
}%
}




\IEEEtitleabstractindextext{
\begin{abstract}
\justifying
Human pose and shape estimation (HPS) has attracted increasing attention in recent years.
While most existing studies focus on HPS from 2D images or videos with inherent depth ambiguity, there are surging need to investigate HPS from 3D point clouds as depth sensors have been frequently employed in commercial devices.
However, real-world sensory 3D points are usually noisy and incomplete, and also human bodies could have different poses of high diversity.
To tackle these challenges, we propose a principled framework, \textbf{\name}, for accurate 3D HPS from point clouds captured in real-world settings, which iteratively refines point features through a cascaded architecture.
Specifically, each stage of \name performs a series of downsampling and upsampling operations to extract and collate both local and global cues, which are further enhanced by two novel modules: \textbf{1)} Cross-stage Feature Fusion (CFF) for multi-scale feature propagation that allows information to flow effectively through the stages, and \textbf{2)} Intermediate Feature Enhancement (IFE) for body-aware feature aggregation that improves feature quality after each stage. 
Notably, previous benchmarks for HPS from point clouds consist of synthetic data with over-simplified settings (\eg, SURREAL) or real data with limited diversity (\eg, MHAD).
To facilitate a comprehensive study under various scenarios, we conduct our experiments on two large-scale benchmarks, comprising \textbf{i)} a dataset that features diverse subjects and actions captured by real commercial sensors in a laboratory environment, and \textbf{ii)} controlled synthetic data generated with realistic considerations such as clothed humans in crowded outdoor scenes. 
Extensive experiments demonstrate that \name, with its powerful point feature extraction and processing scheme, outperforms State-of-the-Art methods by significant margins across the board. 
Ablation studies validate the effectiveness of the cascaded architecture, powered by CFF and IFE.
The pretrained models, code, and data will be publicly available to facilitate future investigation in HPS from point clouds. Homepage: \url{https://caizhongang.github.io/projects/PointHPS/}.

\end{abstract}

}

\maketitle
\IEEEdisplaynontitleabstractindextext
\IEEEpeerreviewmaketitle

\section{Introduction}
\begin{figure*}[t]
  \centering
  \includegraphics[width=\linewidth]{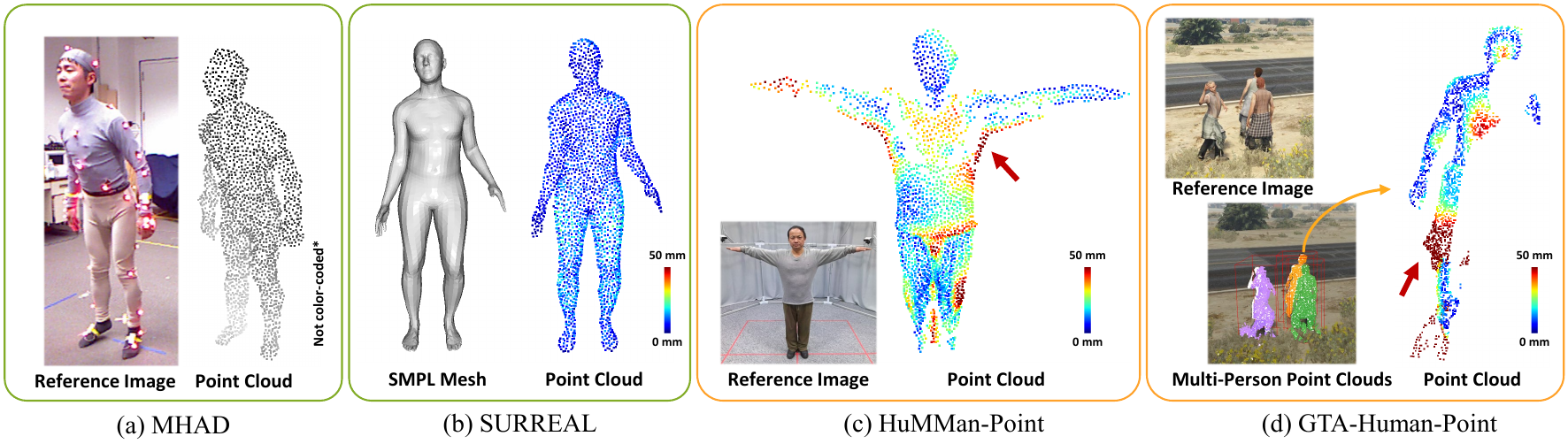}
  \setlength{\abovecaptionskip}{-4mm}
  \caption{Benchmarks for HPS from point clouds. Conventional benchmarks include a) MHAD~\cite{ofli2013berkeley} and b) SURREAL~\cite{Varol2017LearningFS}. Additionally, we include two large-scale benchmarks with realistic considerations: c) HuMMan-Point with real-captured point clouds, and d) GTA-Human-Point with controlled synthetic multi-person scenes (the severely occluded subject's point cloud is shown in detail). The point color indicates the distance to the nearest ground truth SMPL vertex, except for MHAD which has no SMPL annotation. Note that SURREAL directly samples points from unclothed SMPL mesh surfaces~\cite{liu2021votehmr}. However, red arrows in c) and d) show garments induce significant gaps between observed points and the underlying human body. Detailed attributes are summarized in \Tab~\ref{tab:benchmarks}.}
  \label{fig:teaser}
\end{figure*}


Recovering 3D human bodies requires the estimation of both human poses and shapes, which is crucial for many applications, such as physically plausible motion capture, virtual try-on, and mixed reality.
With the development of parametric human models (\eg, SMPL~\cite{loper2015smpl}), human pose and shape estimation (HPS) could be resolved by predicting the shape and pose parameters of parametric models.
Significant progress has been made in predicting SMPL models from 2D images in recent years~\cite{kanazawa2018end, kolotouros2019learning, kocabas2021pare, li2022cliff}.
However, 2D cameras suffer inherent depth ambiguities, giving rise to difficulties in precise distance localization or dimension measurement.
In addition, it also raises concerns to capture human images in privacy-critical environments.
Fortunately, the development of depth sensors and the availability of large-scale datasets \cite{cai2022humman} have opened up a promising avenue for HPS from 3D points, which holds the potential to mitigate many issues of 2D images, hence enabling accurate 3D human recovery.

\begin{table}
  \setlength{\abovecaptionskip}{0cm}    
  \caption{Comparison between similar research tracks that take point clouds as the input and produce human parametric model. As LiDAR-based MoCap focuses on recovering human motions (pose sequences) from sparse point clouds, we identify our track of study (HPS from Point Cloud) most similar to Human Reconstruction, for which both pose and shape can be obtained. Optimization-based post-processing usually lowers inference speed (FPS: frames per second) significantly. *: FPS is not reported \cite{li2022lidarcap, dai2022hsc4d}.}
  \centering
  \small
  \begin{tabular}{lccccc}
    \toprule
    Track & Pose & Shape & Direct & FPS \\
    \midrule
    LiDAR MoCap \cite{li2022lidarcap, dai2022hsc4d} 
        & $\checkmark$ & $\times$ & $\checkmark$ & * \\
    Human Recon. \cite{bhatnagar2020combining, wang2021locally} 
        & $\checkmark$ & $\checkmark$ & $\times$ & $<$0.01 \\ 
    HPS from P.C. \cite{jiang2019skeleton, liu2021votehmr} 
        & $\checkmark$ & $\checkmark$ & $\checkmark$ & $>$25 \\
    \bottomrule
  \end{tabular}
  \label{tab:track_comparison}
\end{table}

However, estimating parametric humans with diverse poses and shapes from real-captured 3D point clouds is far from a trivial task. 
Despite its significance, limited research has been conducted in this area. 
In \Tab\ref{tab:track_comparison}, we categorize the relevant prior studies into 3 groups: 
1) LiDAR-based motion capture methods~\cite{li2022lidarcap, dai2022hsc4d, zhao2022lidar} focus on human pose sequence estimation and typically simplifies shape estimation (\eg, using a mean shape).
2) 3D human mesh reconstruction methods, such as IP-Net~\cite{bhatnagar2020combining} and PTF~\cite{wang2021locally}, firstly generate 3D mesh models based on the input point clouds and then register parametric human models (\eg, SMPL) by using offline optimization, which usually takes a few minutes to reconstruct a high-quality 3D human model for well-prepared clean human points.
3) Parametric human model prediction methods estimate both human body shapes and poses simultaneously from the input 3D point clouds by directly regressing human shape and pose parameters.

This work falls into the third category.
Unlike existing studies~\cite{shi2019skeleton, liu2021votehmr} use synthetic data with simplified assumptions (\eg, sampling noise-free point clouds on the surface of unclothed subjects in SURREAL~\cite{Varol2017LearningFS} as the input), we take a step further and utilize point clouds captured under various realistic settings for recovering 3D human bodies with different shapes and poses, which could directly facilitate many real-applications that require to rapidly resolve HPS based on raw-captured 3D points.
It is noteworthy that HPS from real 3D point clouds is challenging because real-captured 3D human points are often imperfect due to many factors, such as noisy observations from commercial depth sensors (\eg, Microsoft Kinect, ToF sensors on mobile phones), incomplete human shapes due to occlusion in crowded scenes, and also flexible garments frequently causing large deformation of point distribution.
Consequently, it is crucial to study directly estimating human parametric models from 3D point clouds that are captured under realistic scenarios.

%
%
%
%

Inspired by the success of cascaded networks for human pose understanding in 2D vision~\cite{newell2016stacked, chen2018cascaded, chen2019hybrid, sun2019hrnet}, we present \name, a cascaded hierarchical network that repetitively refines and enhances features for human pose and shape estimation from point clouds. 
Designed with a goal to extract essential information from imperfect input, especially when regional cues are inadequate, \name builds upon the point hierarchical architecture~\cite{qi2017pointnet++} by incorporating repeated downsampling and upsampling processes, leading to a powerful feature extraction mechanism with enhanced receptive field~\cite{chen2018cascaded}. 
Interestingly, our experiments show that simply stacking feature extractors does not necessarily bring performance improvements: there are missing key components in a \textit{naive} cascaded architecture that are critical to its effectiveness.  
To address this, we emphasize the design of cross-stage feature passageways and the utilization of intermediate estimation results.
We introduce Cross-stage Feature Fusion (CFF) modules, which densely propagate features from the previous to the current stage in the cascaded architecture. Moreover, CFF adaptively fuses features from different stages, significantly improving parametric model estimation. 
We also design Intermediate Feature Enhancement (IFE) modules to enhance the human shape and pose features based on intermediate parametric model estimations from the early stages. Specifically, IFE groups neighboring point features based on the human joint positions of the intermediate estimations to adaptively enhance human body awareness of encoded features for a more accurate final estimation.

{
\begin{table}
  \setlength{\tabcolsep}{3.5pt}
  \setlength{\abovecaptionskip}{0cm}
  \caption{
  Benchmark statistics. HuMMan-Point consists of three orders of magnitude more frames than existing real benchmark MHAD. SURREAL does not consider garments. GTA-Human-Point features multi-person scenes. Syn.: Synthetic. M.P.: Multi-person scenes. Subj.: Subjects. Act.: Actions.
  }
  \small
  \centering
  \begin{tabular}{l|cccccc}
    \toprule
    \fontsize{7.75}{7.75}\selectfont Dataset & 
    \fontsize{7.75}{7.75}\selectfont Type & 
    \fontsize{7.75}{7.75}\selectfont Clothed & 
    \fontsize{7.75}{7.75}\selectfont M.P. & 
    \fontsize{7.75}{7.75}\selectfont \#Subj. & 
    \fontsize{7.75}{7.75}\selectfont \#Act. & 
    \fontsize{7.75}{7.75}\selectfont \#Frames \\
    \midrule
    \fontsize{7.25}{7.25}\selectfont MHAD & 
    \fontsize{7}{7}\selectfont Real & 
    \fontsize{7}{7}\selectfont Yes &
    \fontsize{7}{7}\selectfont No & 
    \fontsize{7}{7}\selectfont 12 & 
    \fontsize{7}{7}\selectfont 11 & \fontsize{7}{7}\selectfont 2.4K \\
    
    \fontsize{7.25}{7.25}\selectfont SURREAL & 
    \fontsize{7}{7}\selectfont Syn. & 
    \fontsize{7}{7}\selectfont No & 
    \fontsize{7}{7}\selectfont No & 
    \fontsize{7}{7}\selectfont 145 & 
    \fontsize{7}{7}\selectfont N.A. & 
    \fontsize{7}{7}\selectfont 1M \\
    
    \midrule
    
    \fontsize{7.25}{7.25}\selectfont HuMMan-Point & 
    \fontsize{7}{7}\selectfont Real & 
    \fontsize{7}{7}\selectfont Yes & 
    \fontsize{7}{7}\selectfont No & 
    \fontsize{7}{7}\selectfont 1000 & 
    \fontsize{7}{7}\selectfont 500 & 
    \fontsize{7}{7}\selectfont 2.4M \\
    
    \fontsize{7.25}{7.25}\selectfont GTA-Human-Point & 
    \fontsize{7}{7}\selectfont Syn. & 
    \fontsize{7}{7}\selectfont Yes &
    \fontsize{7}{7}\selectfont Yes & 
    \fontsize{7}{7}\selectfont 600 & 
    \fontsize{7}{7}\selectfont 20K & 
    \fontsize{7}{7}\selectfont 750K \\
    \bottomrule
  \end{tabular}
  \label{tab:benchmarks}
\end{table}
}
Furthermore, to facilitate a large-scale evaluation of methods under a realistic setting, we include two new benchmarks, HuMMan-Point and GTA-Human-Point, in addition to conventional benchmarks such as SURREAL~\cite{Varol2017LearningFS} and MHAD~\cite{ofli2013berkeley}. 
In Fig.~\ref{fig:teaser}, we compare and showcase the four benchmarks.
For the two conventional benchmarks, MHAD is a small, real-captured human dataset without SMPL annotations, and SURREAL is a synthetic dataset with point clouds directly sampled from SMPL mesh surfaces.
In contrast, HuMMan-Point and GTA-Human-Point provide large-scale point clouds under a more realistic and relevant setting.
For HuMMan-Point, we adapt the recent large-scale human dataset, HuMMan \cite{cai2022humman}, that provides real-captured point cloud data paired with SMPL annotations of vastly diverse subjects and actions. 
For GTA-Human-Point, we extend the toolchain of GTA-Human \cite{cai2021playing} to generate a large-scale dataset with multi-human scenes that include complex occlusion cases that mimic realistic observation of human group activities.
The detailed comparisons among those benchmark datasets are reported in Table~\ref{tab:benchmarks}.
%
Extensive experiments on various datasets, including HuMMan-Point and GTA-Human-Point, SURREAL, and MHAD, demonstrate that \name surpasses existing state-of-the-art (SoTA) methods by convincing margins across the board. In addition, the ablation study validates our design choices in formulating the cascaded architecture with critical modules IFE and CFF. 

Our contributions are summarized:
\textbf{First}, we propose \name, a novel method with a cascaded architecture designed to enhance point feature extraction from challenging point clouds. We develop modules for comprehensive cross-stage feature fusion and human prior-based refinement that are critical to boosting the effectiveness of the cascaded architecture.
\textbf{Second}, we conduct an initial large-scale study on realistic point cloud-based human pose and shape estimation. Our proposed \name achieves significant performance improvements over existing arts in all evaluated metrics and datasets (notably large-scale real-captured dataset HuMMan-Point and realistic synthetic dataset GTA-Human-Point). We hope our baselines and benchmarks could facilitate future research. 

\begin{figure*}[]
  \centering
  \includegraphics[width=0.90\linewidth]{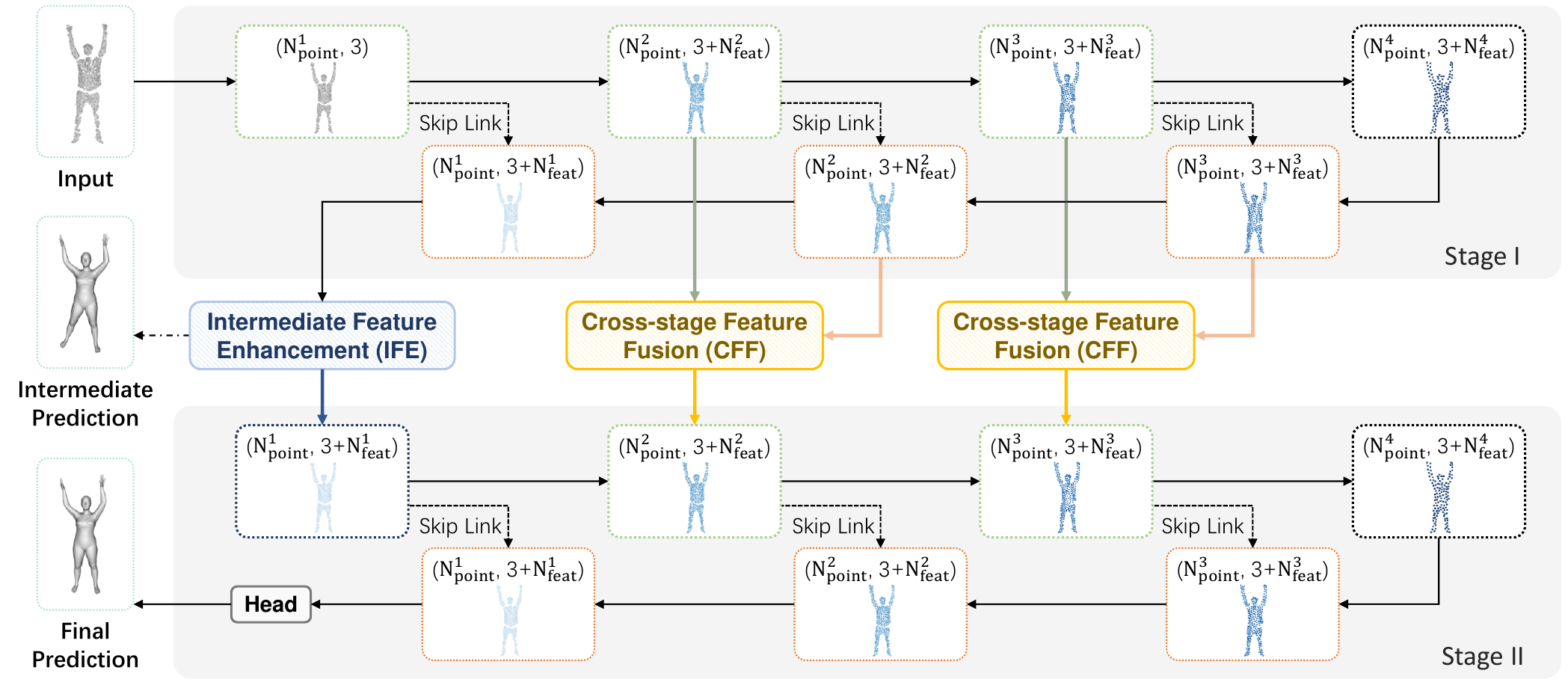}
  \setlength{\abovecaptionskip}{1mm}
  \caption{A two-stage variant of \name is depicted, which is also used as the default in the experiments. \name exhibits a cascaded architecture for iterative feature refinement (\Sec~\ref{sec:method:cascaded}) in consecutive stages. In each stage, there are downsampling (green dotted) and upsampling (orange dotted) modules. The Cross-stage Feature Fusion (CFF, \Sec~\ref{sec:method:cff}) module is illustrated in \Fig\ref{fig:method_cff}. The Intermediate Feature Enhancement (IFE, \Sec~\ref{sec:method:ife}) module is illustrated in \Fig\ref{fig:method_ife}. Details of the downsampling/upsampling modules, and the head are in the \Supp.}
  \label{fig:method}
\end{figure*}

\section{Related Work}
\label{sec:related_work}

\noindent\textbf{Human Pose and Shape Estimation from 2D Input.}
To represent natural human bodies with compact representations, parametric models (\eg, SMPL \cite{loper2015smpl}, SMPL-X \cite{pavlakos2019expressive}, STAR \cite{osman2020star} and GHUM \cite{xu2020ghum}) that are learned from thousands of 3D body scans have been proposed in recent years.
Based on the parametric models, SMPLify \cite{bogo2016keep} estimates 3D body shapes and poses from 2D joints by minimizing keypoint reprojection error, and the optimization process typically takes minutes.
HMR~\cite{kanazawa2018end} introduces an end-to-end learning-based framework to directly regress SMPL parameters from a single RGB image.
Following similar regression schemes, many following works design advanced network modules to predict human models from 2D images~\cite{pavlakos2018learning, omran2018neural, guler2019holopose, kolotouros2019learning, kolotouros2019convolutional, li2020hybrik, georgakis2020hierarchical, li2020hybrik, luo20203d, sun2020monocular, kocabas2021pare, dwivedi2021learning,  kolotouros2021probabilistic, li2022cliff, rajasegaran2022tracking, sun2022putting, wang2023zolly} or monocular videos~\cite{kanazawa2019learning, sun2019human, mehta2020xnect, moon2020i2l, luo20203d, kocabas2020vibe, choi2021beyond}.
However, the 2D images have limitations such as the inherent depth ambiguity leads to difficulties in 3D localization and measurement, and the rich appearance information leads to privacy concerns.  

\noindent\textbf{Human Pose and Shape Estimation from 3D Input.} 
Point cloud has become increasingly popular with the fast development of various 3D sensors, such as low-power LiDAR on mobile devices and consumer-level depth sensors.
Estimating 3D human poses from point clouds (or depth maps) \cite{shotton2011real, haque2016towards, xiong2019a2j, zhou2020learning, garau2021deca} is a long-standing research topic. 
Recently, estimating SMPL models provides a more holistic description of the human body by including the body shape information in the parametric mesh model. 
%
%
To this end, Bashirov \etal~\cite{bashirov2021real} uses an efficient forward kinematic optimization operation to register Kinect skeletons to a fixed parametric human model.
A few methods~\cite{li2022lidarcap, dai2022hsc4d, zhao2022lidar} leverage LiDAR point clouds and SMPL representation for human pose estimation, which could achieve long-range motion capture.
However, those methods~\cite{shotton2011real, haque2016towards, garau2021deca, bashirov2021real, li2022lidarcap, dai2022hsc4d, zhao2022lidar} overlook human shape reconstruction, which is largely attributed to the sparsity of LiDAR point clouds, especially at a far distance.
Instead of directly predicting parametric models, IPNet~\cite{bhatnagar2020combining} and PTF~\cite{wang2021locally} first perform implicit reconstruction to provide body shape and pose initialization, based on which they could register SMPL models on meshes reconstructed from point clouds.
Nonetheless, those methods~\cite{bhatnagar2020combining, wang2021locally} mainly focus on offline human body reconstruction.
There is also a line of methods~\cite{jiang2019skeleton, wang2020sequential, liu2021votehmr} that focus on directly generating parametric human models from point clouds.
Amongst them, Jiang \etal~\cite{jiang2019skeleton} propose the first work that estimates SMPL models from 3D point clouds, but they simplify the 3D point clouds input by directly sampling them from SMPL surfaces.
%
VoteHMR~\cite{liu2021votehmr}, leverages a voting module to aggregate joint features from point features and achieves the best performance to date on synthetic benchmarks. 
However, they~\cite{jiang2019skeleton, liu2021votehmr} are evaluated mainly on synthetic datasets with simplified settings.
In this work, we focus on processing realistic point clouds (\eg, clothed humans captured by commercial depth sensors) for parametric human recovery.

\section{\name}
\label{sec:method}

\noindent
\textbf{Overview.}
In this study, we strive to recover parametric human models (\ie, SMPL) from partially-observed point clouds of clothed humans under various natural poses. 
We follow the mainstream top-down paradigm \cite{kanazawa2018end, kolotouros2019learning, liu2021votehmr} in this study: we first use 3D bounding boxes to crop single-person point clouds from the observation as our network input, and then the proposed \name regresses the SMPL parameters for each person individually.

\noindent
\textbf{SMPL.} 
Skinned Multi-Person Linear model (SMPL) is a popular parametric human body model that employs a low-dimensional vector $\beta \in \mathbb{R}^{10}$ to control body shapes, a vector $\theta \in \mathbb{R}^{24\times 3}$ to represent poses in axis-angle representation. 
To model a specific human body, SMPL starts with a mean-shape template and applies blend shape functions to represent body shape and dynamic soft-tissue deformations caused by pose deviations. 
The model also regresses joint locations from shape parameters and incorporates bone transformations to generate a posed and shaped mesh of the human body. 
Generally, the SMPL model can be formally defined as $J,V=M_{\textrm{SMPL}}(\beta, \theta)$, which generates 3D human mesh of vertices $V \in \mathbb{R}^{6890 \times 3}$, and 3D joints $J \in \mathbb{R}^{24 \times 3}$ of various human shapes and poses.

\subsection{Cascaded Architecture}
\label{sec:method:cascaded}

Encouraged by the success of cascaded networks in 2D vision \cite{newell2016stacked, chen2018cascaded, chen2019hybrid, sun2019hrnet}, we extend the point hierarchical architecture~\cite{qi2017pointnet++} to a cascaded framework by stacking multiple \textit{stages}, which could achieve an iterative feature refinement~\cite{newell2016stacked} for human model estimation.
Here we define a \textit{stage} as a series of downsampling and upsampling modules. They perform multi-scale feature extraction, which captures both fine-grain details at the local level and collective cues at the global level. 
We show a two-stage \name in \Fig~\ref{fig:method}. In a \textit{plain} cascaded architecture, the first stage takes point cloud 3D positions as the input, and the following stage then takes the point cloud positions and processed features from the previous stage to further refine the features. $N^{n}_{point}$ and $N^{n}_{feat}$ are the number of points and point feature dimension at each scale $n$. 
However, we discover that directly concatenating two backbones does not necessarily lead to improvements. To address this problem, we further build \name, with Cross-stage Feature Fusion (CFF) in \Sec~\ref{sec:method:cff} and Intermediate Feature Enhancement (IFE) in \Sec~\ref{sec:method:ife} in our cascaded model. 
%
Below we provide a brief introduction to the upsampling/downsampling modules and the regression head, and their details can be found in the \Supp.

\noindent \textbf{Upsampling and Downsampling Modules.}
Unlike images with pixels arranged in an orderly lattice, point clouds are unordered, sparse, and irregular \cite{qi2017pointnet, wang2019dynamic, qin2020bipointnet}. Hence, to achieve similar functions as image upsampling and downsampling layers, we utilize the popular Set Abstraction and Feature Propagation modules \cite{qi2017pointnet++}. 

\noindent \textbf{Head for Parameter Regression.}
As SMPL consists of pose parameters, $\theta \in \mathbb{R}^{24 \times 3}$, and shape parameters, $\beta \in \mathbb{R}^{10}$, we adapt the head from HMR \cite{kanazawa2018end} in our work to regress these parameters. Moreover, we remove the weak-perspective camera parameter regression because the 2D keypoint reprojection error is not available for the loss computation. 

\begin{figure}
    \centering
    \includegraphics[width=0.9\linewidth]{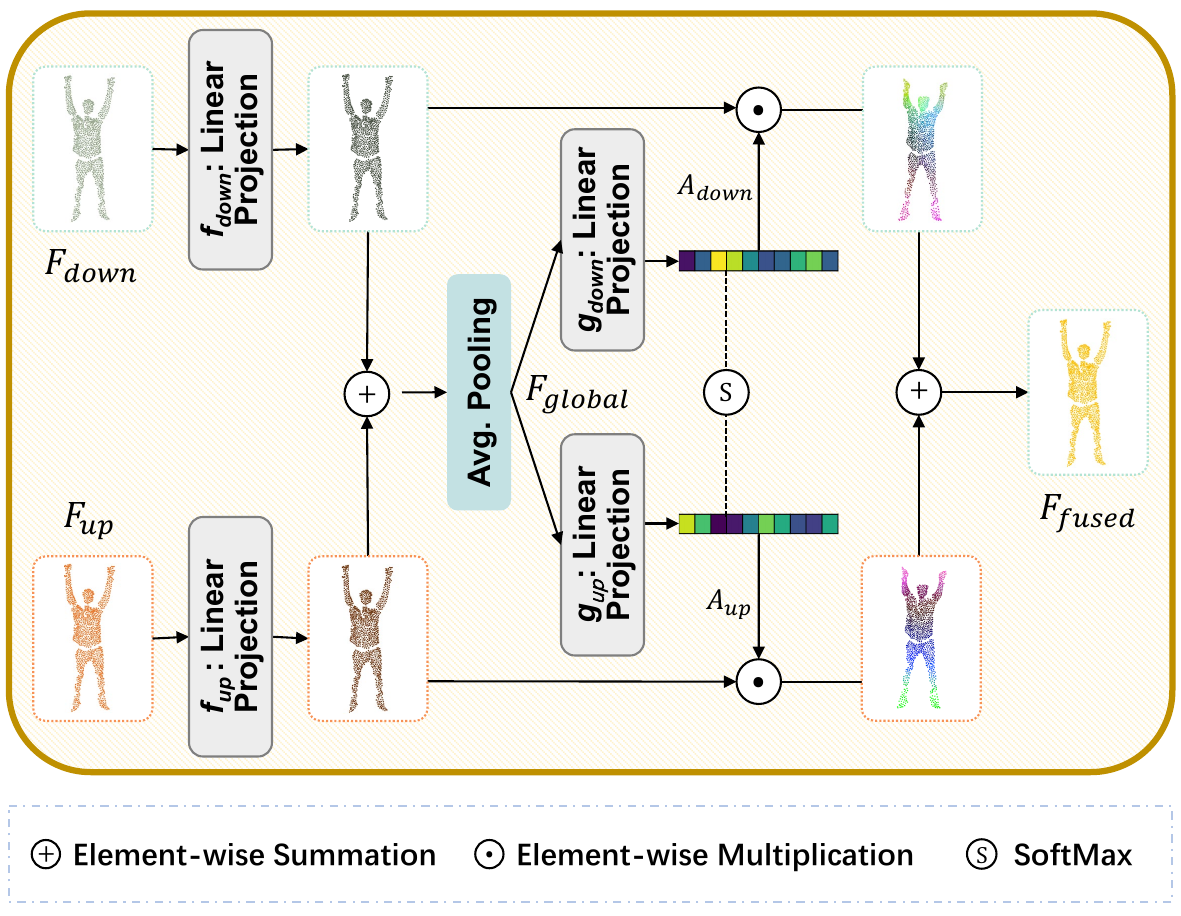}
    \vspace{-2.5mm}
    \caption{Cross-stage Feature Fusion (CFF, \Sec~\ref{sec:method:cff}) enables feature to propagate at multiple scales for feature fusion in an adaptive manner.}
    \label{fig:method_cff}
\end{figure}
\begin{figure*}[]
  \centering
  \includegraphics[width=\linewidth]{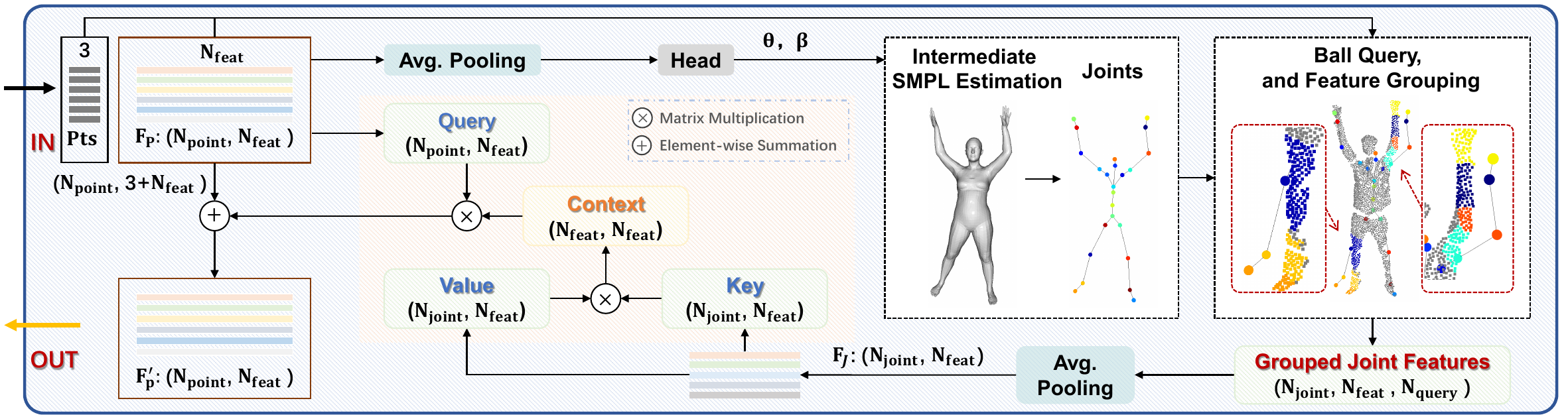}
  \setlength{\abovecaptionskip}{0cm}
  \vspace{-3mm}
  \caption{Iterative Feature Enhancement (IFE, \Sec~\ref{sec:method:ife}) produces intermediate SMPL estimation between feature extraction stages in the cascaded architecture. It leverages the intermediate 3D joint positions as anchor points to query and aggregate neighboring features. The pooled features enhance the original feature through an efficient attention mechanism. }
  \label{fig:method_ife}
\end{figure*}
\subsection{Cross-stage Feature Fusion}
\label{sec:method:cff}

To encourage interactions between stages in the cascaded architecture, we introduce the Cross-stage Feature Fusion (CFF) module as feature passageways at multiple scales in the hierarchy to propagate features from previous stages to the current stage (shown in \Fig~\ref{fig:method}).
CFF takes in features from both the downsampling phase and upsampling phase to enrich the feature aggregated during multi-scale feature extraction, which receives benefits from different levels of semantics. Furthermore, features from different phases may carry complementary information. Hence, we explore various fusion strategies, such as adaptive fusion that dynamically balances the weights of features with different semantics. Inspired by Selective Kernel \cite{li2019selective}, we design CFF to adaptively select input features of different semantic levels. In contrast to Selective Kernel which employs convolution layers of various kernel sizes on images, CFF leverages existing point features at the same scale but different phases in the feature extraction stage to achieve a large performance gain.

Particularly, the features from the downsampling phase and the upsampling phase of the same scale are first projected via MLP layers, and then undergo selective weighted feature fusion shown in Fig.~\ref{fig:method_cff}. 
A compact feature descriptor $F_{global}$ is first computed with features from both upsampling phase ($F_{up}$) and downsampling phase ($F_{down}$):
\begin{equation}
    F_{global} = p(f_{up}(F_{up}) + f_{down}(F_{down}))    
\end{equation}
where $p$ is global average pooling, $f$ is a set of linear projection.
With $N_{feat}$ as the number of feature channels, the adaptive weights $A_{up}, A_{down} \in \mathbb{R}^{N_{feat}}$ are computed by channel-wise SoftMax ($\sigma$) operation:
\begin{equation}
    A^{c}_{up}, A^{c}_{down} = \sigma([g_{up}(F_{global})]^{c}, [g_{down}(F_{global})]^{c})
\end{equation}
where $g$ is another set of linear projections, $\sigma$ is Softmax operation. Note that $c$ indicates the operation is applied by channel. The adaptive feature fusion is performed:
\begin{equation}
    F_{fused} = A_{up} \cdot f_{up}(F_{up}) + A_{down} \cdot f_{down}(F_{down})
\end{equation}
Our experimental results show that the CFF modules are an indispensable component for \name to achieve high-performing 3D parametric human recovery.

\subsection{Intermediate Feature Enhancement}
\label{sec:method:ife}

To further enhance the feature refinement in the cascaded scheme, we introduce the Intermediate Feature Enhancement (IFE) module (shown in Fig.~\ref{fig:method_ife}).
Specifically, the IFE module first estimates the SMPL parameters after a stage as the intermediate results. Afterward, we reconstruct the SMPL body and regress the joints. We then use the human joints as anchor points to group neighboring point features ($N_{feat}$ feature channels) with ball query operations using a radius $r$, and the joint features $F_{J} \in \mathbb{R}^{N_{J}\times N_{feat}}$ are encoded with an average pooling over the grouped point features $F_{query} \in \mathbb{R}^{N_{J}\times N_{feat} \times N_{query}}$.

Subsequently, we utilize the Efficient Attention \cite{shen2021efficient} to enhance the point features with the joint features:
\begin{equation}
    F_{P}' = \sigma_{q}(f_{q}(F_{P}))\sigma_{k}((f_{k}(F_{J}))^{T}f_{v}(F_{J})),
\end{equation}
where $f$ and $\sigma$ are linear projection and SoftMax layers, and $F_{J}$, $F_{P}$, and $F_{P}'$ are joint features, point features, and enhanced point features, respectively. 
Note that Efficient Attention approximates the standard Transformer Decoder \cite{vaswani2017attention} that reduces cross-attention to linear complexity, which significantly reduces the memory footprint and is critical when the number of points is large.

\subsection{Loss Functions}
\label{sec:method:loss}
We penalize the predicted human poses in rotation representation $R\in \mathbb{R}^{3\times3}$ and the predicted shapes $\beta$ with loss functions $\mathcal{L}_{\theta}$ and $\mathcal{L}_{\beta}$, respectively.
Formally, they are defined as:
\begin{equation}
    \centering
    \mathcal{L}_{\theta} = \frac{1}{N_{j}}\sum_{i}^{N_{j}} ||R_{i} - \hat{R_{i}}||_{2}, \;\;\; \mathcal{L}_{\beta} = ||\beta - \hat{\beta}||_{2},
\end{equation}
\noindent
where $N_{j}$ is the number of joints, and $\hat{R_{i}}$ and $\hat{\beta}$ are the corresponding ground truth rotation and shape parameters, respectively.
In addition, L1 loss is used for penalizing per-joint ($\mathcal{L}_{j}$) and per-vertex ($\mathcal{L}_{v}$) errors.
\begin{equation}
  \mathcal{L}_{j} = \frac{1}{N_{j}}\sum_{j_{i} \in J, \hat{j_{i}} \in \hat{J}} |j_{i} - \hat{j_{i}}|_{1},  
\end{equation}
\begin{equation}
  \mathcal{L}_{v} = \frac{1}{N_{v}}\sum_{v_{i} \in V, \hat{v_{i}} \in \hat{V}} |v_{i} - \hat{v_{i}}|_{1},  
\end{equation}
\noindent
where joints $J$ and vertices $V$ are obtained from SMPL model $M_{\textrm{SMPL}}$ with pose and shape parameters $J, V = M_{\textrm{SMPL}}(\beta, \theta)$.
Consequently, the total loss is defined as:
\begin{equation}
  \mathcal{L} = \lambda_{j}\mathcal{L}_{j} + \lambda_{v}\mathcal{L}_{v} + \sum_{i}^{N_{s}}( \lambda_{\theta, i}\mathcal{L}_{\theta} + \lambda_{\beta, i}\mathcal{L}_{\beta}),
\end{equation}
where $N_{s}$ is the number of stages, and $\lambda$ represents the corresponding loss weights. 
Note for intermediate SMPL estimations, only parameter losses are used to reduce the number of hyperparameters.
\begin{figure*}[t]
  \centering
  \includegraphics[width=\linewidth]{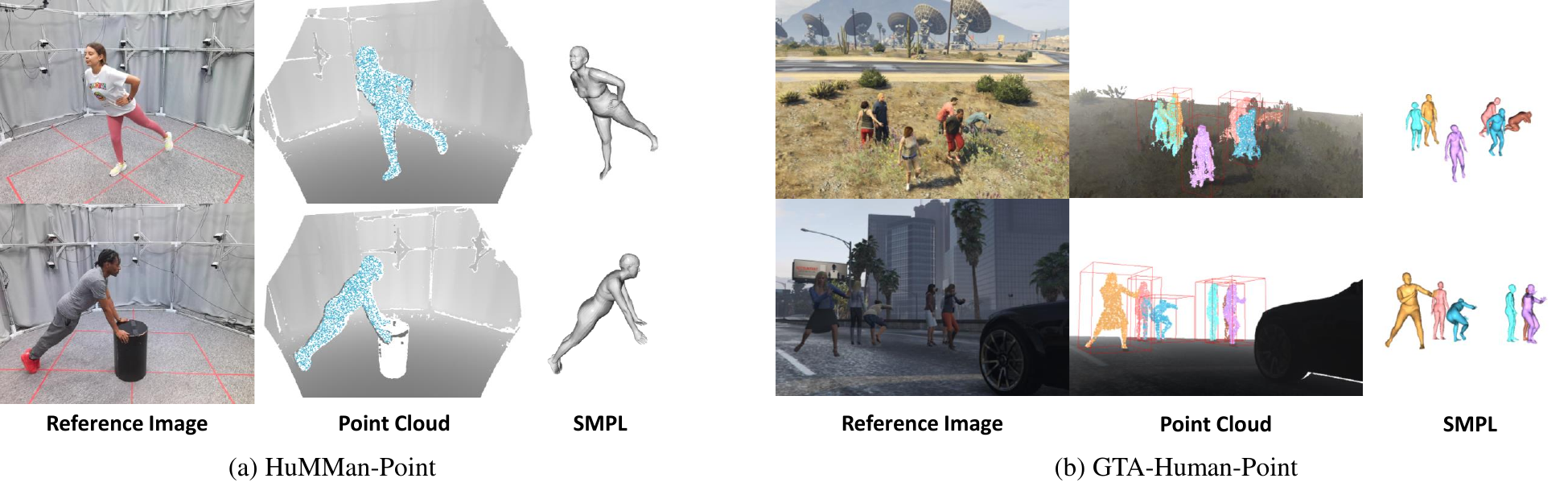}
  \setlength{\abovecaptionskip}{-5mm}
  \caption{We include two large-scale benchmarks (HuMMan-Point and GTA-Human-Point) in our study. a) HuMMan-Point is a real-captured, large-scale dataset with diverse subjects and poses. b) GTA-Human-Point features occlusions (\eg, top right) and realistic subject-scene interaction (\eg, the bottom right example shows subjects reacting to an approaching vehicle). More visualization is included in the \Supp.}
  \label{fig:benchmarks}
\end{figure*}
\section{Experiments}
\label{sec:experiments}
We conduct extensive experiments on various datasets such as HuMMan-Point, GTA-Human-Point, SURREAL, and MHAD. Moreover, we conduct an ablation study to verify the contribution of the proposed architecture and modules. In addition, we visualize key results, and discuss efficiency and occlusion.

\subsection{Evaluation Metrics}

We use standard metrics for human pose and shape estimation. The primary metric is PVE, the mean L2 distance between the 6890 ground truth and predicted SMPL mesh vertices, which takes into consideration both the pose and the shape. MPJPE measures L2 error for joints regressed from SMPL mesh. It reflects mainly the pose quality, including the global orientation. PA-MPJPE computes per-joint errors after Procrutes Alignment, and it focuses on body pose quality. Besides, we follow VoteHMR \cite{liu2021votehmr} to use uni-directional Chamfer distance (CD) instead of PVE as the metric on MHAD, which does not provide ground truth SMPL annotations. All metrics are in millimeters (mm).

\subsection{Evaluation Benchmarks}

We conduct extensive experiments on two new, large-scale benchmarks (HuMMan-Point and GTA-Human-Point) and two standard benchmarks (SURREAL and MHAD). In \Tab~\ref{tab:benchmarks}, key attributes of the four benchmarks are compared. 

\noindent \textbf{HuMMan-Point.}
HuMMan \cite{cai2022humman} is a recent mega-scale human dataset with depth images captured by Kinects. Its full set contains 1000 subjects, 500 actions, 400 thousand clips (40 thousand video sequences, each having 10 synchronized Kinect views) and 60 million frames. Note that HuMMan is a multi-view dataset, but we use monocular videos independently in our study. \Fig~\ref{fig:benchmarks}a) showcases a variety of subjects and poses. However, it is not practical to use all views due to its prohibitive scale. Hence, we downsample a subset of HuMMan by sampling from 10 Kinect views while considering subject and pose diversity. The details of the sampling strategy is included in the \Supp. The derived dataset (named HuMMan-Point) contains 11,000 training sequences and 5,000 test sequences, giving rise to around 2.4 million frames of data in total. 

\noindent \textbf{GTA-Human-Point.}
GTA-Human \cite{cai2021playing} is a large-scale dataset built with the popular video game Grand Theft Auto with accurate SMPL labels. It features diverse subjects and actions, with realistic human-scene interaction. We extend the original data generation toolchain in two ways. First, we intercept the rendering pipeline to obtain the depth maps, which we convert to point clouds. More details are included in the \Supp. Note that we truncate the point clouds more than 10 m away from the camera as the typical maximum range of commercial depth sensors does not exceed 10 m. Second, we create multi-person scenarios. These scenarios feature realistic interaction between subjects and environment as shown in \Fig~\ref{fig:benchmarks}b). We randomize in-game location selection, subject placements and action assignments, leading to complicated scenes with noisy point clouds (\eg, from the grass) and occlusions. The new dataset is named GTA-Human-Point and it comprises 20K sequences, with each scene containing 2-6 subjects. These sequences are randomly split into training and test set at a ratio of 7:3. Despite that GTA-Human-Point is a synthetic dataset, we highlight that the data are different from SURREAL because subjects are clothed, and there is more realistic interaction between subjects and the environment. 

\noindent \textbf{SURREAL \cite{Varol2017LearningFS}.}
We follow VoteHMR \cite{liu2021votehmr} and Want \etal \cite{wang2020sequential} to place SMPL models from SURREAL dataset and render the depth maps. We sample 10 thousand sequences (referred to as ``clips" in SURREAL), which comprise around one million depth images. The test split consists of 10 thousand depth images and the rest are used for training. The depth images are converted to point clouds with intrinsic parameters of the virtual cameras used in the rendering process. SURREAL is a large-scale dataset. However, note that the generated point clouds are sampled directly from SMPL model surfaces, which neglects the impact of garments. There is no subject-scene interaction either.

\noindent \textbf{MHAD \cite{ofli2013berkeley}.} 
It is a small-scale datasets (11 actions and 12 subjects) captured with two Kinect sensors. We follow the official instruction to convert depth images to point clouds. We also follow VoteHMR \cite{liu2021votehmr} to sample 2400 frames from the first eight sequences of both Kinects as the test set.

\begin{figure*}[t]
  \centering
  \includegraphics[width=0.9\linewidth]{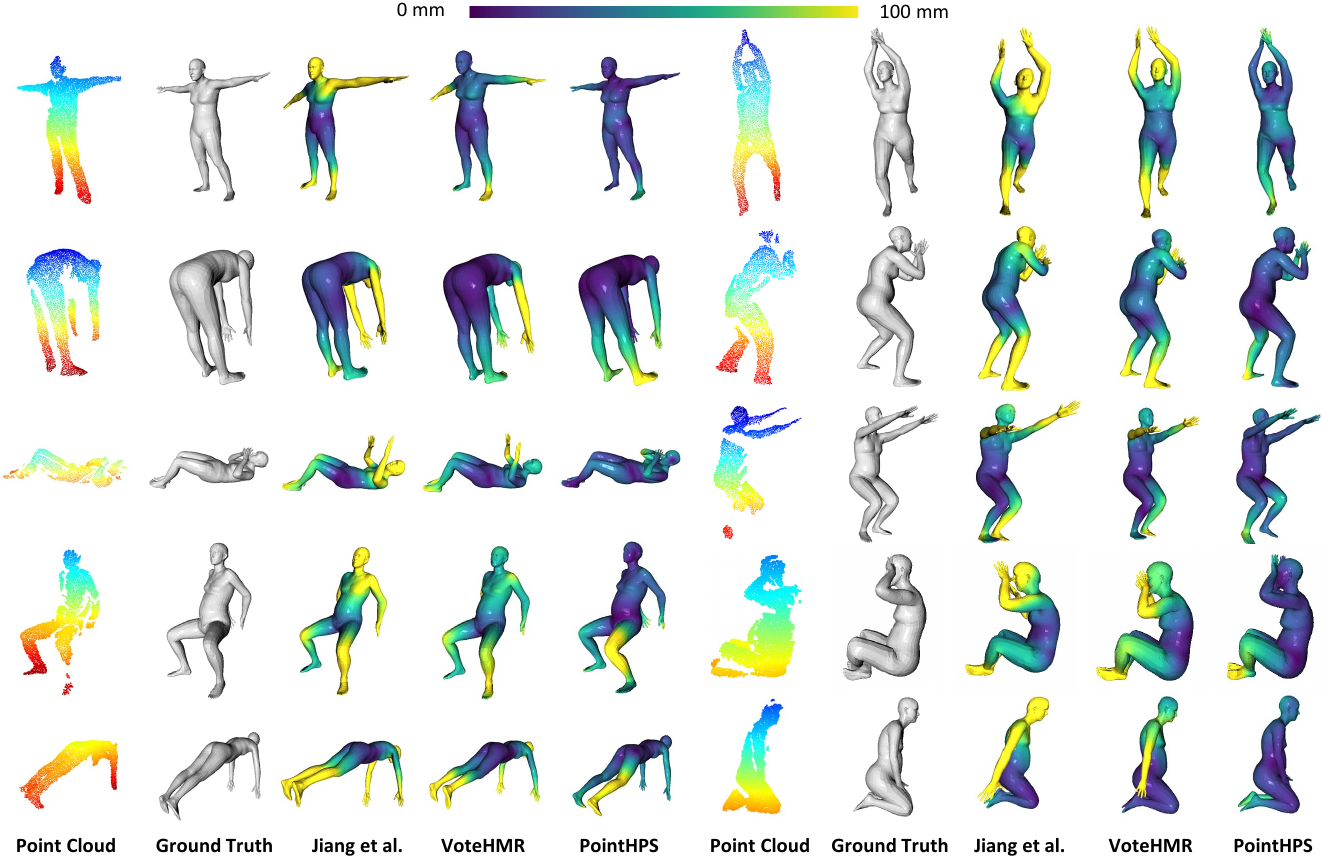}
  \setlength{\abovecaptionskip}{1mm}
  \caption{Visualization of results on HuMMan-Point. We observe that \name is able to handle difficult input point clouds given clothed subjects with various poses, noisy measurements, and missing patches. The color on the predicted SMPL mesh indicates vertex-to-vertex error whereas the color of the point cloud is for visualization purposes only (this is the same for \Fig\ref{fig:visual_compare_gta},\ref{fig:visual_iphone}, and \ref{fig:visual_compare_ptf}).}
  \label{fig:visual_compare}
\end{figure*}
\subsection{Implementation Details}
We follow the standard top-down paradigm in human pose and shape estimation \cite{kanazawa2018end, kolotouros2019learning, liu2021votehmr} and crop single-person point clouds with ground truth 3D bounding boxes. We then recenter them with 3D bounding box center coordinates, and downsample each point cloud to 2048 points as the input to the models. No scaling is applied to avoid disrupting human shape estimation. 
For Cascaded Architecture, increasing the number of stages leads to a greater memory footprint, computation cost, and difficulty in training. Hence, we evaluate \name up to three stages in this work. In each stage, we follow PointNet++ \cite{qi2017pointnet++} to use five levels of scales. $N_{point} = \{2048, 1024, 512, 256, 128\}$ and $N_{feat} = 128$ for all levels. For Intermediate Feature Enhancement, we use $r_{query}$ = 0.2 m and $N_{query}$ = 64. Ball query and feature grouping are similar to that used in the downsampling module. For loss weights, we use $\lambda_{j} = 100$ and $\lambda_{v} = 2$. For \name, the last stage with the primary head is unchanged, whereas IFE is used in other stages with secondary heads. We find that the loss weights of the secondary heads should be kept low. Otherwise, they would disrupt the training of the primary head. Hence, we use $\lambda_{\theta} = \{1.5, 3\}$ and $\lambda_{\beta} = \{0.01, 0.02\}$ for the two-stage variant and $\lambda_{\theta} = \{0.6, 0.9, 3\}$ and $\lambda_{\beta} = \{0.004, 0.006, 0.02\}$ for the three-stage variant.
All training is conducted on 8 V100 GPUs, the batch size is 1024 for all baselines and \name with two stages. Adam optimizer is used with a learning rate of 0.0025 for 60 epochs for all baselines on HuMMan-Point, and GTA-Human-Point. We follow VoteHMR~\cite{liu2021votehmr} for to use learning rate of 0.0001 on SURREAL. No weight decay or data augmentation is used in all experiments.

\subsection{Results on Large-scale Benchmarks}
We conduct experiments on HuMMan-Point and GTA-Human-Point Benchmark to gauge the model performances in realistic situations. HuMMan-Point is a large-scale dataset with diverse actions, making it a comprehensive benchmark for human-centric analysis. As for GTA-Human-Point, it provides realistically simulated incomplete point clouds due to occlusions. We train two milestone works to compare with \name: Jiang \etal\cite{jiang2019skeleton} is the pioneering deep learning-based method for estimation of SMPL parameters from point clouds; VoteHMR \cite{liu2021votehmr} is the SoTA method.

\noindent \textbf{HuMMan-Point.}
In \Tab~\ref{tab:humman}, we show experiment results on HuMMan-Point. HuMMan-Point provides a better representation of real-life environments due to its large scale and rich collection of diverse subjects and actions. \name is able to achieve impressive performance boosts over existing methods, with more than 10 mm improvements on all three key metrics: PVE, MPJPE, and PA-MPJPE. 

\begin{table}
  \setlength{\abovecaptionskip}{0cm}
  \caption{Results on HuMMan-Point. \name outperforms existing methods by convincing margins across all metrics. The unit for all metrics is millimeters (mm).}
  \centering
  \small
  \begin{tabular}{l|ccc}
    \toprule
    Method & PVE $\downarrow$ & MPJPE $\downarrow$ & PA-MPJPE $\downarrow$ \\
    \midrule
    Jiang \etal \cite{jiang2019skeleton} & 147.5 & 124.1 & 102.3 \\
    VoteHMR \cite{liu2021votehmr} & 97.6 & 81.1 & 63.4 \\
    \midrule
    \name & \textbf{81.9} & \textbf{70.1} & \textbf{52.0} \\
    \bottomrule
  \end{tabular}
  \label{tab:humman}
\end{table}

\begin{table}
  \setlength{\abovecaptionskip}{0cm}
  \caption{Results on GTA-Human-Point. \name achieves significant improvements over existing baselines. The unit for all metrics is millimeters (mm).}
  \small
  \centering
  \begin{tabular}{l|ccc}
    \toprule
    Method & PVE $\downarrow$ & MPJPE $\downarrow$ & PA-MPJPE $\downarrow$ \\
    \midrule
    Jiang \etal \cite{jiang2019skeleton} & 163.1 & 135.9 & 110.8 \\
    VoteHMR \cite{liu2021votehmr} & 114.4 & 95.7 & 75.0 \\
    \midrule
    \name & \textbf{107.0} & \textbf{92.9} & \textbf{69.1} \\
    \bottomrule
  \end{tabular}
  
  \label{tab:gta-human++}
\end{table}

\noindent \textbf{GTA-Human-Point.}
In \Tab\ref{tab:gta-human++}, we show experiment results on GTA-Human-Point. Compared to HuMMan-Point, GTA-Human-Point consists of scenes with relatively more extensive occlusion, and the performance of all methods degrades. Nevertheless, with cascaded feature refinement, fusion, and enhancement, \name is able to capture global cues to supplement for the missing local cues. As a result, \name still surpasses other methods by convincing margins.

\begin{figure*}[t]
  \centering
  \includegraphics[width=0.9\linewidth]{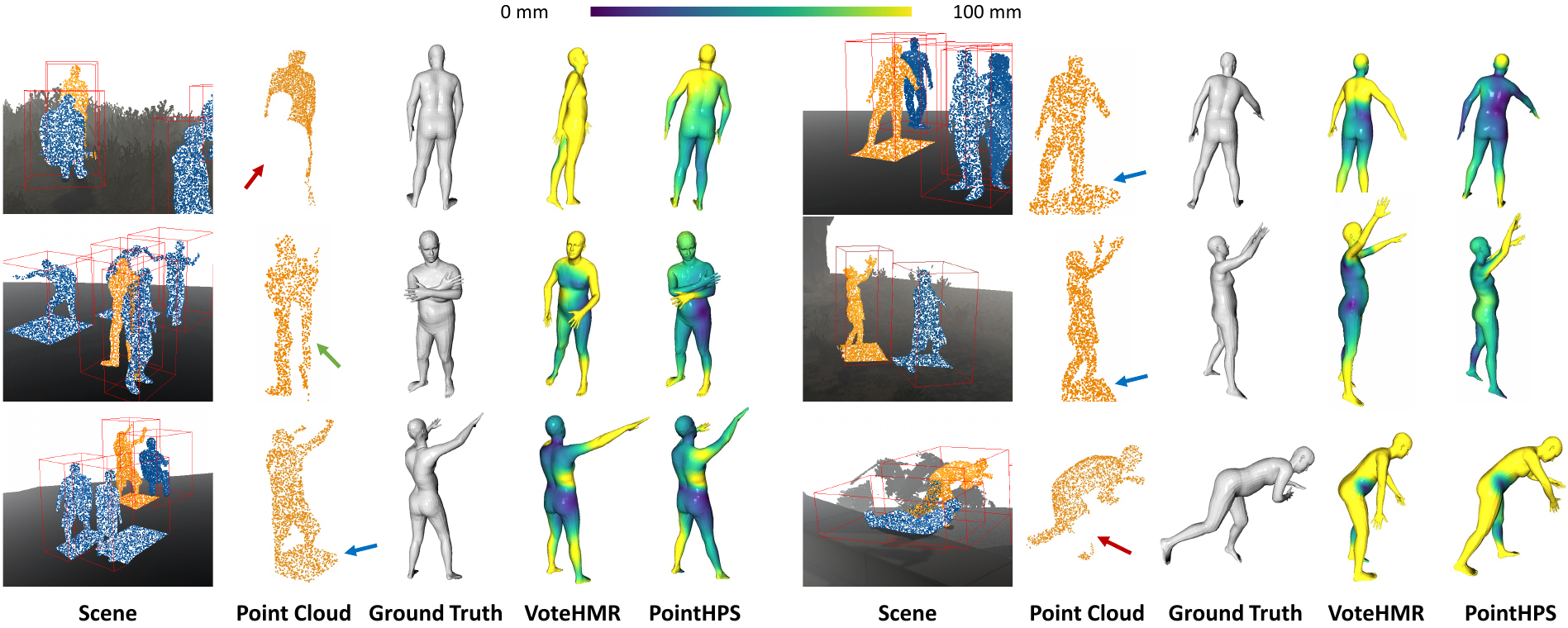}
  \setlength{\abovecaptionskip}{1mm}
  \caption{Visualization of \name's performance on GTA-Human-Point, which features scenes with occlusion, giving rise to incomplete human point clouds. We show the estimation results of the subjects marked with orange points. Incomplete point clouds and rare poses remain difficult to solve, especially when they co-exist, but \name still gives reasonable predictions. In addition, challenging parts of the point clouds are indicated by arrows. Red arrow: incompleteness. Green arrow: intrusion of other subjects in proximity. Blue arrow: environment or background. }
  \label{fig:visual_compare_gta}
\end{figure*}
\noindent \textbf{Qualitative Comparisons}
\label{sec:exp:visualization}
We include visualization for a comparison of different baselines in \Fig\ref{fig:visual_compare}. \name produces favorable results and is relatively more robust to various body poses and incomplete point clouds. In addition, we compare \name with VoteHMR on GTA-Human-Point, which consists of scenes with multiple subjects in close proximity, giving rise to severe occlusion in \Fig\ref{fig:visual_compare_gta}. \name is less prone to missing points and thus handles incomplete point clouds relatively better than VoteHMR.

\begin{figure}[t]
  \centering
  \includegraphics[width=\linewidth]{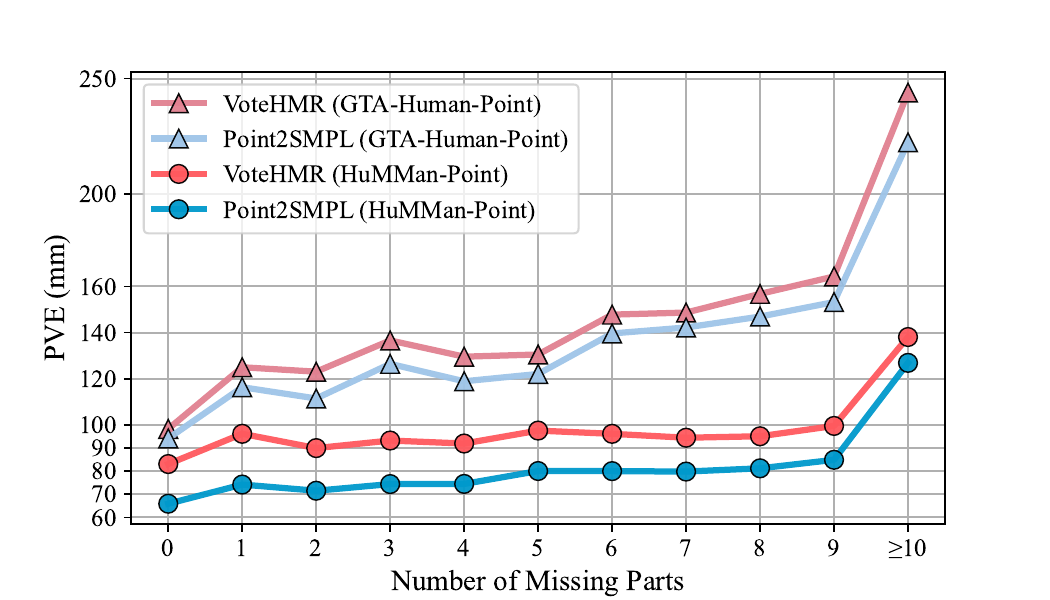}
  \setlength{\abovecaptionskip}{-3mm}
  \caption{Model performance against a number of missing parts. More missing parts indicate more severe occlusion (including self-occlusion) and lead to performance degradation. We group ten or more missing parts together for there are much fewer examples. }
  \label{fig:pve_vs_visibility_all}
\end{figure}
\noindent \textbf{Occlusion Analysis.}
It is common to have incomplete point clouds due to occlusions. In Figure \ref{fig:pve_vs_visibility_all}, we represent occlusion by the number of missing body parts based on SMPL's body segmentation of 24 body parts. Occlusion causes performance degradation in all methods, but \name with a cascaded architecture consistently outperforms VoteHMR, which uses a joint completion module to handle missing parts. Moreover, the gap between GTA-Human-Point and HuMMan-Point may be attributed to several factors that make point clouds more challenging in GTA-Human-Point: 1) intrusion of points that belong to other subjects nearby and 2) intrusion of environment or background points. Some examples are shown in \Fig\ref{fig:visual_compare_gta}. Note that in practice, background is easily removed in a static studio setup (\eg, HuMMan-Point) but not trivial in the wild.

\begin{table}
  \setlength{\abovecaptionskip}{0cm}
  \caption{Results on the standard benchmarks SURREAL and MHAD. \name achieves the best performance. Since MHAD does not provide SMPL annotations, CD (uni-directional Chamfer distance) is used as the metric.}
  \label{tab:surreal-mhad}
  \centering
  \small
  \begin{tabular}{l|cc}
    \toprule
      & SURREAL & MHAD \\
    \cmidrule{2-2} \cmidrule{3-3}
    Method & PVE (mm) $\downarrow$ & CD (mm) $\downarrow$ \\
    \midrule
    Jiang \etal \cite{jiang2019skeleton} & 80.5 & 121.6 \\   
    3DCODED \cite{groueix20183d} & 41.8 & 56.2 \\
    Wang \etal \cite{wang2020sequential} & 24.3 & 81.2 \\
    VoteHMR \cite{liu2021votehmr} & 20.2 & 51.7 \\
    \midrule
    \name & \textbf{19.6} & \textbf{48.5} \\
    \bottomrule
  \end{tabular}
\end{table}
\subsection{Results on Conventional Benchmarks}
\begin{figure*}[t]
  \centering
  \includegraphics[width=0.85\linewidth]{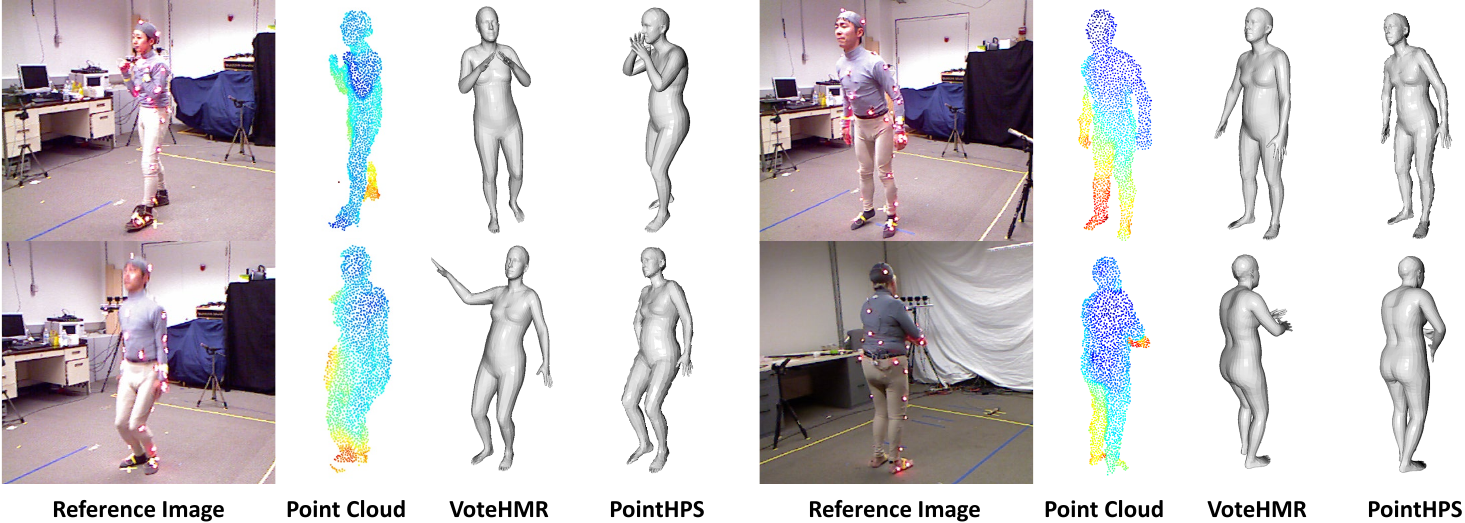}
  \setlength{\abovecaptionskip}{0cm}
  \vspace{2mm}
  \caption{Visualization of results on MHAD. \name handles different body poses well, even without training on MHAD. The color of the point cloud is for visualization purposes only. MHAD does not provide SMPL annotations.}
  \label{fig:visual_compare_mhad}
\end{figure*}

We conduct experiments on SURREAL and MHAD in \Tab\ref{tab:surreal-mhad}, where \name achieves the SoTA performance. Note that models are directly used to run inference on MHAD after pretraining on SURREAL, without fine-tuning, following the same protocol as that in VoteHMR \cite{liu2021votehmr}. The performance gains on SURREAL and MHAD are relatively smaller than that on HuMMan-Point and GTA-Human-Point. We conjecture that this is because the SURREAL holds a simplified assumption of on-surface points, and MHAD is small in scale and diversity, resulting in relatively easy performance saturation. Note that the results on the standard benchmarks are obtained without any post-processing, as typical optimization-based post-processing takes seconds, even minutes to converge \cite{wang2021locally, bhatnagar2020combining} and is unsuitable for real-time applications. In addition, we visualize model performance on MHAD in \Fig~\ref{fig:visual_compare_mhad}: \name outperforms VoteHMR when subjected to various poses and noisy point clouds.

\subsection{Visualizations on iPhone Data}
\begin{figure}[t]
  \centering
  \includegraphics[width=\linewidth]{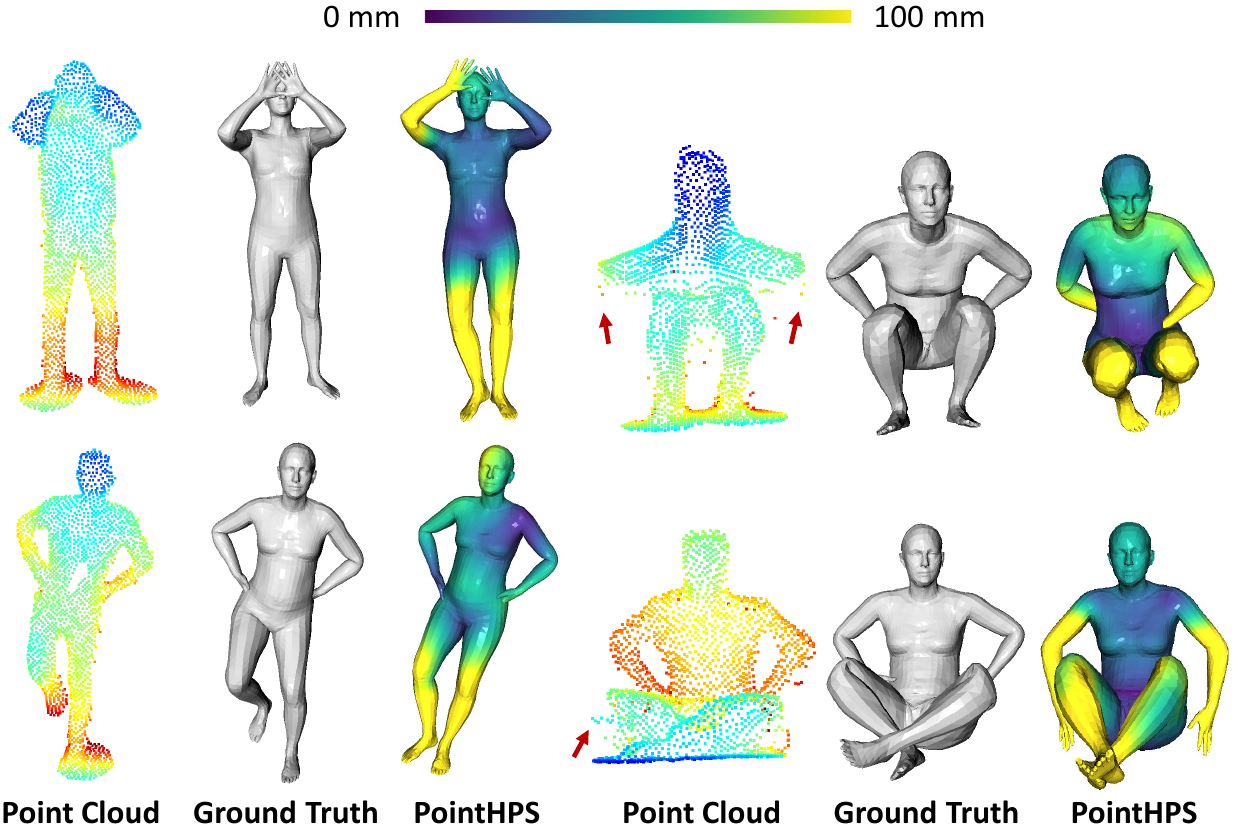}
  \setlength{\abovecaptionskip}{-1mm}
  \caption{Visualization of inference results on iPhone point clouds (typically noisy at the edges, indicated by red arrows). Each triplet consists of point cloud, ground truth SMPL, and estimated SMPL.}
  \label{fig:visual_iphone}
\end{figure}
HuMMan also provides point clouds captured by iPhone. We thus investigate \name's performance on this popular mobile device in \Fig\ref{fig:visual_iphone}. Two-stage \name is used as the model trained on Kinect data and directly tested on iPhone data. It is observed that the point clouds produced by iPhone are much noisier than that by Kinect, especially at the edges. Nevertheless, \name is still able to obtain reasonably good estimations.

\subsection{Comparison with Human Reconstruction.}
\begin{figure}[t]
  \centering
  \includegraphics[width=0.85\linewidth]{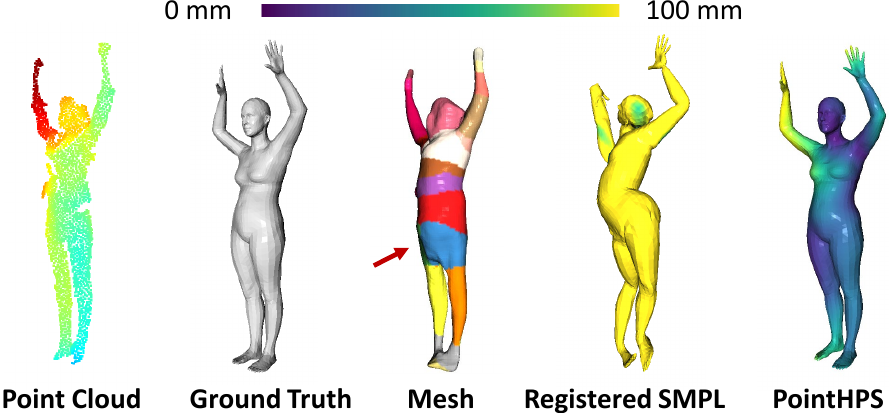}
  \setlength{\abovecaptionskip}{2mm}
  \caption{Comparisons between human reconstruction method with \name. As a representative human mesh reconstruction method, PTF~\cite{wang2021locally} firstly reconstructs body surface meshes (see ``Mesh''), followed by SMPL/SMPL+D registration by optimization (``Registered SMPL''). Imperfect part prediction is indicated by arrow, here we follow PTF's color code: blue is right thigh but both thighs are predicted to be blue. See \Supp for more visualizations.}
  \label{fig:visual_compare_ptf}
\end{figure}

In \Fig~\ref{fig:visual_compare_ptf}, we compare PTF~\cite{wang2021locally} and \name, where the former is a representative work of human reconstruction. This line of work typically reconstructs body surface mesh, and then register SMPL model on it. This paradigm has two disadvantages. First, the quality of SMPL registration relies largely on the mesh reconstruction and part prediction, which are prone to noise and incompleteness of the input point cloud. We find this is a common problem with more visualizations in the \Supp. Second, the registration takes several minutes to converge. Hence, we argue that direct regression may be more suitable for HPS in a real-time context. Note we follow the training setting of PTF~\cite{wang2021locally} and train the pretrained model on HuMMan-Point for a fair comparison.

\subsection{Ablation Study}
\begin{table}
  \setlength{\abovecaptionskip}{0cm}    
  \caption{Ablation study on number of stages in the cascade. Performance improves with an increasing number of stages. One-stage \name is degenerated and cannot have CFF or IFE modules.}
  \centering
  \small
  \begin{tabular}{c|ccc}
    \toprule
    \#Stages & PVE $\downarrow$ & MPJPE $\downarrow$ & PA-MPJPE $\downarrow$ \\
    \midrule
     1 & 111.2 & 94.8 & 70.0 \\
     2 & 81.9 & 70.1 & 52.0 \\ 		
     3 & 79.1 & 68.0 & 50.5 \\
    \bottomrule
  \end{tabular}
  \label{tab:stages}
\end{table}

\begin{table}
  \setlength{\abovecaptionskip}{0cm}    
  \caption{Main ablation study. Interestingly, directly employing multiple stages does not improve performance but IFE and CFF are critical to achieving better performance with more feature extraction stages (\Tab\ref{tab:stages}).  Cascaded: Cascaded Architecture. IFE: Intermediate Feature Enhancement. CFF: Cross-stage Feature Fusion.}
  \centering
  \small
  \begin{tabular}{ccc|ccccc}
    \toprule
    Cascaded & CFF & IFE & PVE $\downarrow$ & MPJPE $\downarrow$ & PA-MPJPE $\downarrow$ \\
    \midrule
     - & - & - & 111.2 & 94.8 & 70.0 \\
     \checkmark & - & - & 113.3 & 97.5 &	71.4 \\
     \checkmark & \checkmark & - & 85.6 & 73.0  &  54.4 \\
     \checkmark & - & \checkmark & 104.8 & 90.0 & 67.1 \\ 	
     \checkmark & \checkmark & \checkmark & 81.9 & 70.1 & 52.0 \\
    \bottomrule
  \end{tabular}
  \label{tab:ablation}
\end{table}
\noindent \textbf{Cascaded Architecture.}
We ablate \name on HuMMan-Point. In \Tab\ref{tab:stages}, we show that the cascaded architecture (equipped with IFE and CFF) is effective: more stages lead to better performance. However, larger models results in more costly computation (\Tab\ref{tab:model_size_infer_speed}). Hence, we conduct other experiments on the two-stage variant of \name. Note that both CFF and IFE cannot work without the cascaded architecture; \name with one stage degenerates to a simplified architecture with a point backbone followed by a head for parameter estimation.

\begin{table}[t]
  \setlength{\abovecaptionskip}{0cm}    
  \caption{Ablation study on Cross-stage Feature Fusion. We find features from both upsampling branch and downsampling branch are beneficial. Adaptive feature fusion further boosts the performance. Up / Down: Features from upsampling / downsampling phase. Adapt: Adaptive feature fusion.}
  \centering
  \small
  \begin{tabular}{ccc|ccc}
    \toprule
    Up & Down & Adapt & PVE $\downarrow$ & MPJPE $\downarrow$ & PA-MPJPE $\downarrow$ \\
    \midrule
     - & - & - & 104.8 & 90.0 & 67.1\\
     \checkmark & - & - & 104.1 & 89.3 & 62.9  \\
     \checkmark & \checkmark & - & 93.1 & 80.1 & 59.5 \\
     \checkmark & \checkmark & \checkmark & 81.9 & 70.1 & 52.0  \\
    \bottomrule
  \end{tabular}
  \label{tab:ablation_cff}
\end{table}

\noindent \textbf{Cross-stage Feature Fusion.} 
We highlight that a high-performing cascaded architecture is much more than an intuitive concatenation of multiple backbones. In \Tab\ref{tab:ablation}, we discover that a naive cascade without CFF or IFE does not achieve better performance. However, with CFF added in, significant performance improvement is observed. 
Furthermore, we conduct a more detailed ablation on CFF in \Tab\ref{tab:ablation_cff}. The results show that aggregation of same-scale features from both the upsampling branch and the downsampling phases are essential, upon which adaptive feature fusion of these features results in the best performance.

\noindent \textbf{Intermediate Feature Enhancement.}
In \Tab\ref{tab:ablation}, adding IFE always leads to performance improvement. This indicates that leveraging intermediate parametric human estimation to enhance features in the cascaded architecture is beneficial. Moreover, IFE is compatible with CFF as combining both modules leads to the best result.

\begin{table}[t]
  \setlength{\abovecaptionskip}{0cm}    
  \caption{Model sizes (number of parameters) and inference speeds (frames per second) of various baselines and variants. The inference is conducted on a V100 GPU. PVE (mm) is obtained on HuMMan-Point.}
  \vspace{1mm}
  \centering
  \small
  \begin{tabular}{l|crrr}
    \toprule
    Method & \#Stages & PVE $\downarrow$ & \#Param $\downarrow$  & FPS $\uparrow$ \\
    \midrule
    Jiang \etal \cite{jiang2019skeleton} & - & 147.5 & 1.44M & 57.8 \\
    VoteHMR \cite{liu2021votehmr}        & - & 97.6 & 20.52M & 74.5 \\
    \midrule
                                    & 1 & 111.2 & 1.84M & 96.3 \\
    \name                           & 2 & 81.9 & 4.70M & 37.4 \\
                                    & 3 & 79.1 & 6.96M & 27.6 \\
    \bottomrule
  \end{tabular}
  \label{tab:model_size_infer_speed}
\end{table}

\subsection{Storage and Computational Efficiency}
In \Tab\ref{tab:model_size_infer_speed}, we evaluate model sizes and inference speed (with a V100 GPU). For \name, having more stages leads to a larger model and additional computation costs. Considering only the real-time methods with $>30$ FPS: Jiang \etal has the smallest model size but is less competitive in terms of error; VoteHMR uses the largest amount of parameters, leading to the largest model size; the two-stage variant of \name is stronger yet more lightweight than VoteHMR.   
\section{Conclusion}
In this work, we propose \name, a robust method for human pose and shape estimation from realistic point clouds. It utilizes a Cascaded Architecture, Cross-stage Feature Fusion, and Intermediate Feature Enhancement modules. We put \name to a large-scale study on realistic point clouds for parametric human recovery, including two comtemporary datasets HuMMan-Point and GTA-Human-Point. Our proposed \name outperforms existing methods convincingly across the board.

\noindent \textbf{Limitations.} 
\name produces SMPL parameters from point clouds. However, simultaneous estimation of more expressive humans with hands and face parameters (\ie, in the form of SMPL-X) remains a challenge.

\ifCLASSOPTIONcompsoc
  \section*{Acknowledgments}
\else
  \section*{Acknowledgment}
\fi

This work is supported by NTU NAP, MOE AcRF Tier 2
(T2EP20221-0033), and under the RIE2020 Industry Alignment Fund - Industry Collaboration Projects (IAF-ICP) Funding Initiative, as well as cash and in-kind contribution from the industry partner(s).

\appendices 


\section{More Implementation Details}
\label{supp:more_implementation_details}

\subsection{Cascaded Architecture}
\begin{figure*}[t]
  \centering
  \includegraphics[width=0.75\linewidth]{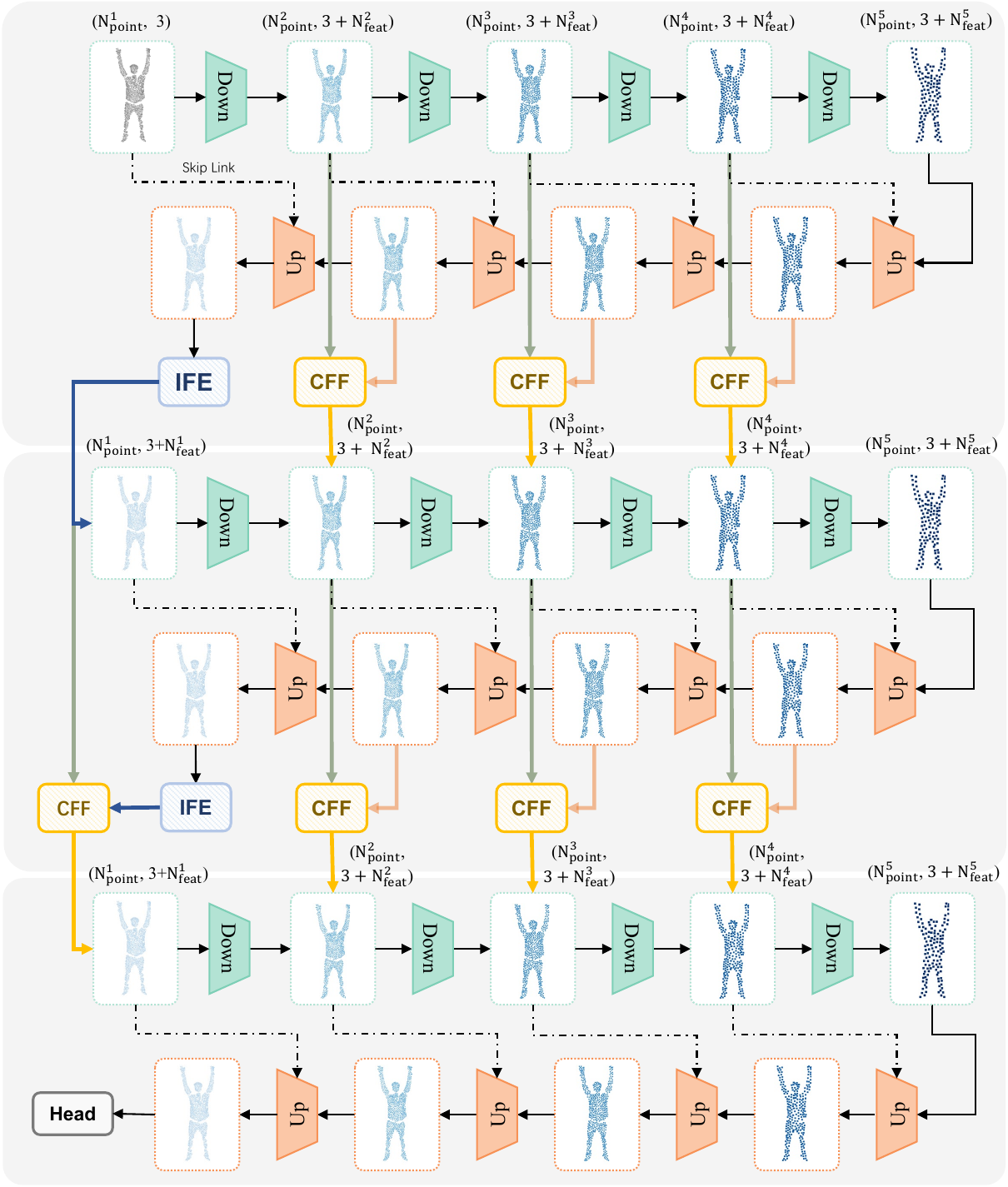}
  \setlength{\abovecaptionskip}{5mm}
  \caption{A three-stage variant of \name showing the cascaded architecture. Detailed illustration of the downsampling and upsampling modules is found in \Fig~\ref{fig:up_and_down}. The input is point clouds with 3 channels (x, y, and z coordinate values). The head (\Fig~\ref{fig:hmr_head}) directly regresses SMPL pose and shape parameters. $N^{n}_{point}$ and $N^{n}_{feat}$ are the number of points and point feature dimension at scale $n$. Dotted lines represent skip links.}
  \label{fig:method_3_stages}
\end{figure*}
For a clear illustration, we draw a variant of \name with three stages of feature refinement in \Fig\ref{fig:method_3_stages}. The second stage might be repeated for a variant with more than three stages. However, having more stages leads to slower computation. Hence, we use the two-stage variant as the default setting in our experiments in the main paper.

\subsection{Upsampling and Downsampling Modules}
\begin{figure*}[t]
  \centering
  \includegraphics[width=\linewidth]{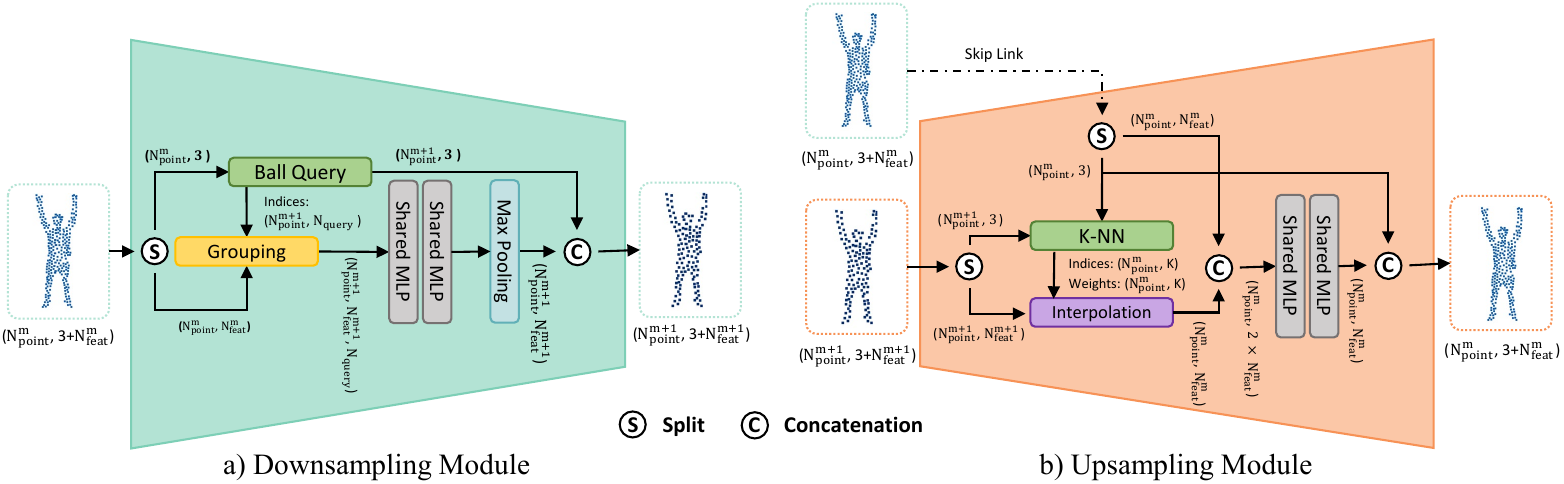}
  \setlength{\abovecaptionskip}{0mm}
  \caption{We illustrate a) a downsampling module and b) a upsampling module at the same scale. The Split operation separates the position channels (three channels for x, y, and z coordinates) and the feature channels. Point clouds in the green boxes are from the downsampling branch whereas point clouds with orange boxes are from the upsampling branch. $N^{m}_{point}$ and $N^{m}_{feat}$ are the number of points and point feature dimension at scale $m$. $N_{query}$ is the maximum number of points queried by a radius. K is the number of nearest neighbors.}
  \label{fig:up_and_down}
\end{figure*}

The upsampling and downsampling modules aim to process and aggregate both local and global cues in the noisy and incomplete point clouds. Note that our cascaded architecture does not require specific designs of such modules; in practice, we follow PointNet++ \cite{qi2017pointnet++} and use its Set Abstraction and Feature Propagation modules for downsampling and upsampling. We illustrate these modules in \Fig~\ref{fig:up_and_down}.

\subsection{Head for Parameter Regression.}
\begin{figure}[t]
  \centering
  \includegraphics[width=0.6\linewidth]{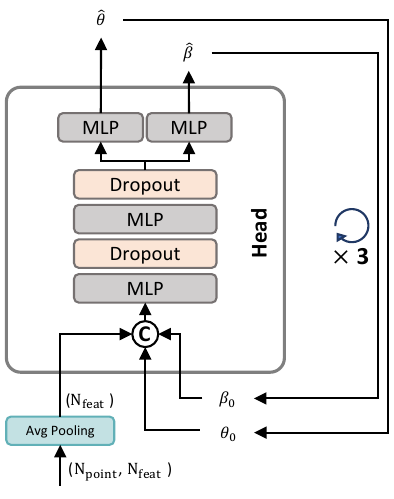}
  \setlength{\abovecaptionskip}{2mm}
  \caption{Modified head for parameter regression. $\theta_{0}$ and $\beta_{0}$ are mean parameter values \cite{kolotouros2019learning} that are used as the initial values in the iterative error feedback \cite{kanazawa2018end} that regress the final parameters through three consecutive estimations. }
  \label{fig:hmr_head}
\end{figure}

The head estimates the body pose $\theta$ and shape $\beta$ parameters. In our implementation, we choose HMR \cite{kanazawa2018end} head with modifications for \name: we remove the weak-perspective camera parameter regression. In 2D HPS, these camera parameters are critical for projecting 3D SMPL joints to 2D image plane, which are then used to compute pixel-space loss against ground truth 2D keypoints. In our study with point cloud as the input, however, we cannot leverage 2D keypoints. Fig~\ref{fig:hmr_head} illustrates the modified head. $\theta_{0}$ and $\beta_{0}$ are mean parameter values \cite{kolotouros2019learning} used to initialize the iterative error feedback \cite{kanazawa2018end}. In each iteration, $\hat{\theta}$ and $\hat{\beta}$ are estimated and used as the initial values for the next iteration. We follow HMR to execute three iterations.

\section{More on Human Reconstruction}
\label{supp:human_recon}
\begin{figure*}[t]
  \centering
  \includegraphics[width=0.9\linewidth]{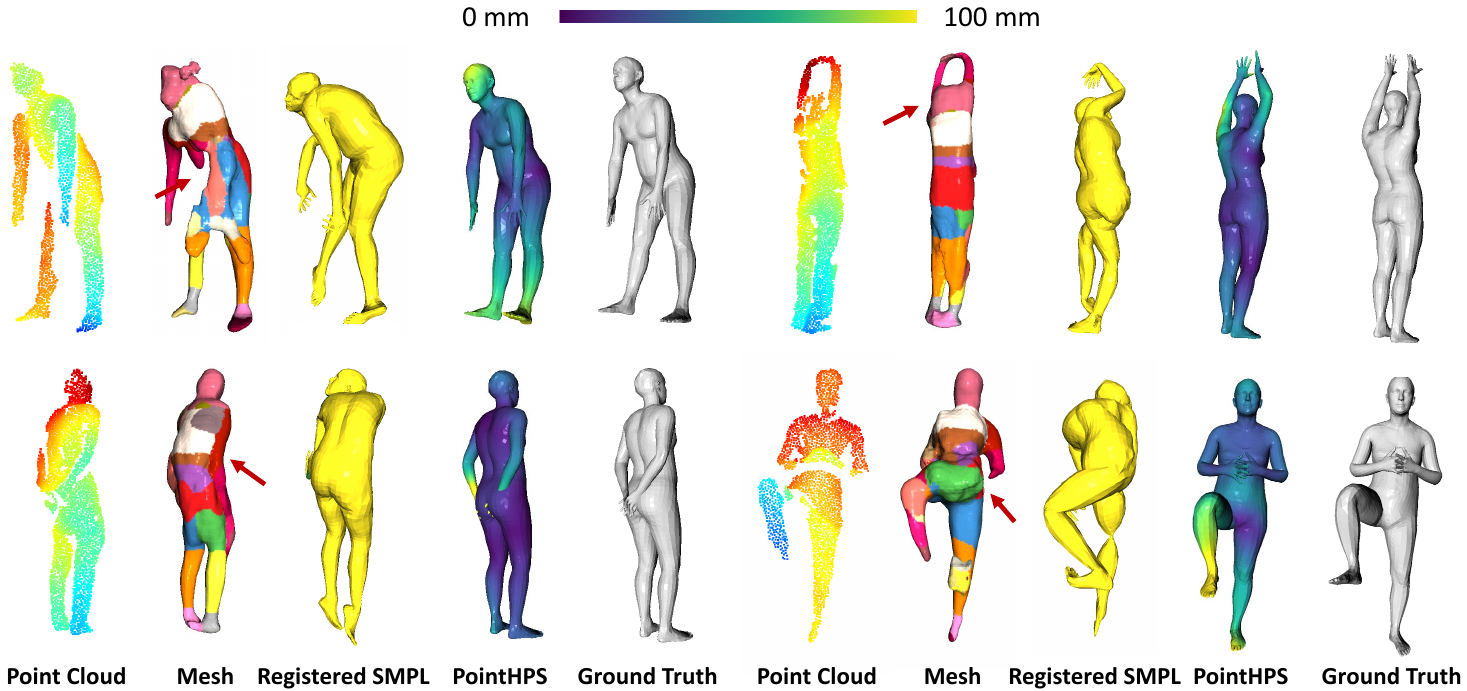}
  \setlength{\abovecaptionskip}{2mm}
  \caption{More comparison between a representative human reconstruction method (PTF~\cite{wang2021locally}) and \name. PTF firstly reconstructs body surface meshes (see ``Mesh''). We follow PTF's color code in rendering the mesh based on the part segmentation prediction. Next, PTF registers SMPL/SMPL+D on the mesh through optimization (see ``Registered SMPL''). The red arrows indicate artifacts in the mesh reconstruction or part segmentation prediction. Top left: there is a missing part in the reconstructed mesh. Top right: the head and the upper arms are all predicted as the ``head". Bottom left: the arms are reconstructed wrongly as they should cross behind the subject's back. Bottom right: the hips are predicted to be the ``right thigh" and the arms are not reconstructed properly.  
  }
  \label{fig:visual_compare_ptf_supp}
\end{figure*}

We include more visualization and analysis on human reconstruction method (PTF~\cite{wang2021locally}) in \Fig~\ref{fig:visual_compare_ptf_supp}. Human reconstruction methods typically reconstructs the mesh first, followed by registration of SMPL parametric model on the mesh surface. We observe that the first step is sensitive to noises and incompleteness in the input point clouds, which is typically the case for point clouds captured by real sensors. Moreover, the second step depends largely on the quality of the reconstructed mesh and the part segmentation results, which tends to be unstable given distorted mesh with erroneous part predictions.

\section{More Details on the Benchmarks}
\label{supp:pointhuman}
\begin{figure*}[b]
  \centering
  \includegraphics[width=\linewidth]{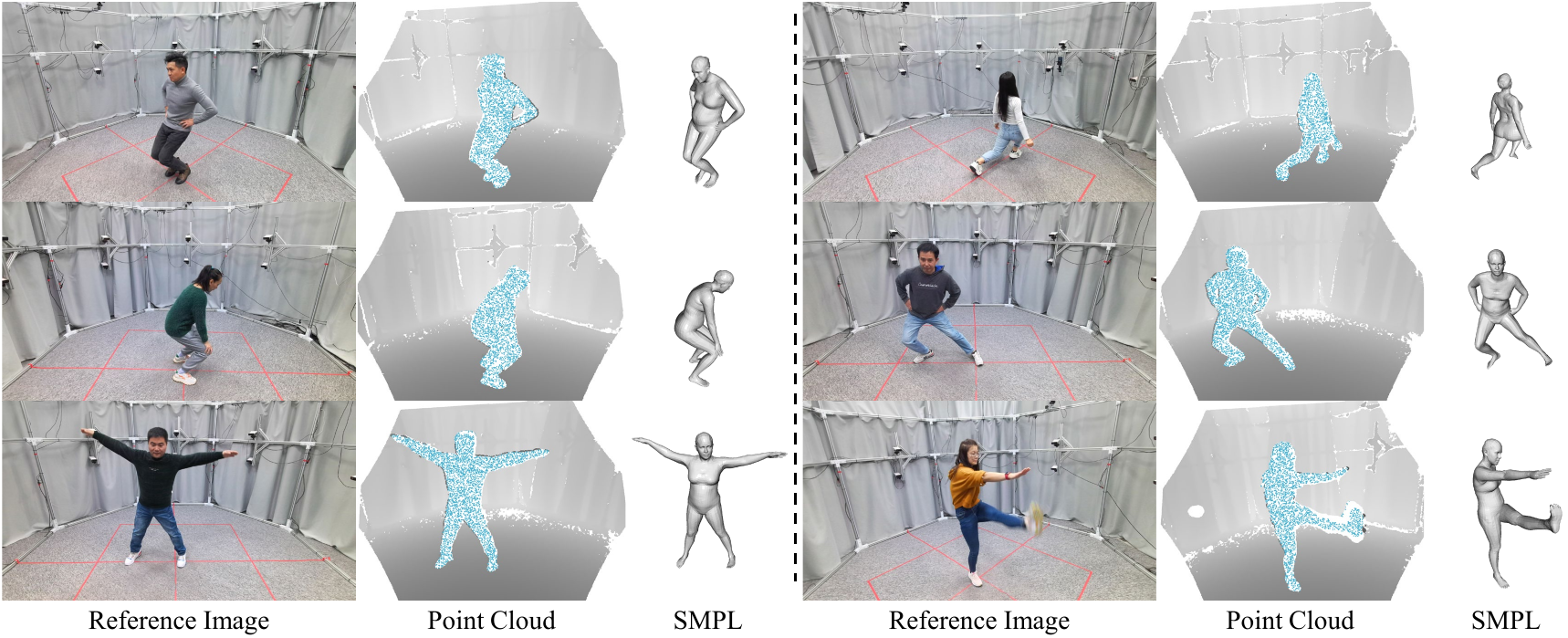}
  \setlength{\abovecaptionskip}{-1mm}
  \caption{More examples from HuMMan-Point dataset, which features diverse subjects and poses. In each example, we show the reference image captured by the RGB camera (for visualization only, images are not used in our study), the point cloud converted from the depth images, that are captured by the depth camera, and the SMPL annotation. For the point cloud visualization, we render the background points in grayscale and the foreground points in blue color.}
  \label{fig:humman_more}
\end{figure*}

To facilitate large-scale study on point cloud-based human pose and shape estimation under a more realistic setting, we introduce HuMMan-Point, which features real-captured point clouds with diverse subjects and poses, and GTA-Human-Point, which consists of multi-person scenes with realistic human-scene interaction and occlusion.

\subsection{HuMMan-Point}
\subsubsection{Data Selection Strategy}
We derive HuMMan-Point from the mega-scale database HuMMan \cite{cai2022humman} by downsampling the original 400 thousand video clips to 16 thousand video clips. Since each monocular video clip has three tags: subject tag (person ID), action tag (action ID), and sensor tag (Kinect ID), a close inspection of the data shows that: 1) different subjects have different numbers of video clips; 2) the number of video clips against the action class is a long-tailed distribution (most frequent action classes have much more video clips than least frequent counterparts); 3) there are equal amount of video clips from each of the 10 Kinect views. Hence, to maintain the diversity and balance in subjects and actions in the downsampled set, we assign weights to each clip: 
\begin{equation}
    w^{i} = \frac{1}{N^{i}_{subj} \times N^{i}_{act}}    
\end{equation}
\noindent where $N^{i}_{subj}$ is the number of video clips with the same subject tag as video clip $i$, and $N^{i}_{act}$ is the number of video clips with the same action tag as video clip $i$. We thus perform weighted sampling without replacement. The probability of video clip ${i}$ being selected is
\begin{equation}
    P(i) = \frac{w^{i}}{\sum^{N}_{j=1}w^{j}}    
\end{equation}
\noindent where $N$ is the number of remaining video clips to select from. The selected video clip is removed from the original set. The selection is performed multiple times until the target number is reached. 

\subsubsection{Foreground Segmentation} 
For each video clip, we use the provided matting masks to crop out the foreground points: we project the point cloud on the image plane and select those points that land on pixels with a non-zero matting value. The foreground points are then downsampled to 2048 points as the input.

\subsubsection{More Visualization} 
We show additional examples from the HuMMan-Point dataset in \Fig~\ref{fig:humman_more}. HuMMan-Point consists of diverse subjects and expressive human poses, making it a comprehensive benchmark for parametric human recovery from real-capture point clouds. 

\subsection{GTA-Human-Point}
\begin{figure*}[t]
  \centering
  \includegraphics[width=\linewidth]{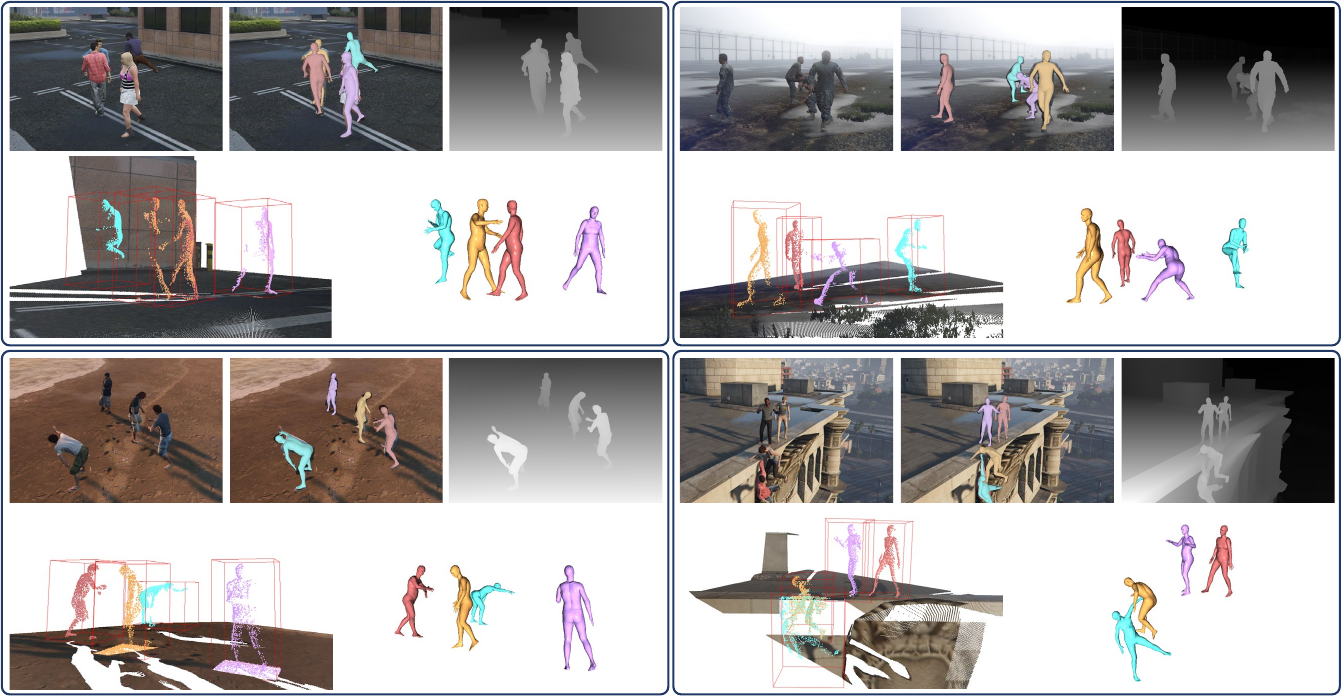}
  \setlength{\abovecaptionskip}{2mm}
  \caption{Examples from GTA-Human-Point dataset. For each example, the top three images are RGB images, SMPL annotations overlaid on RGB images, and depth maps. The bottom two are rendered point clouds with 3D bounding boxes, and corresponding SMPL annotations in the 3D space. Note that the bottom images are rendered from a new view that is different from the original virtual camera for better visualization of incomplete point clouds due to occlusion. Moreover, the color of the SMPL meshes and point clouds indicate correspondences. }
  \label{fig:gta_human++}
\end{figure*}

\subsubsection{Toolchain}
We build our toolchain based on GTA-Human \cite{cai2021playing} and JTA \cite{fabbri2018learning} that produces images paired with SMPL annotations. However, GTA-Human does not provide point clouds. To close this gap, we attach our thread to intercept the DirectX-powered rendering pipeline, and read the Z-buffer (depth buffer) values to obtain pixel-wise raw depth values. However, the raw depth values cannot be used directly; they are converted in the following equation \cite{cao2020long}.

\begin{equation}
    d = \frac{d_{far} * d_{near}}{d_{raw} * (d_{far} - d_{near})}
\end{equation}

\noindent
where $d$ is the real depth value, $d_{raw}$ is the raw depth value, $d_{near} = 0.15$ and $d_{far} = 2000$ are two constants. 
Afterward, point clouds are created from depth maps with known camera intrinsics \cite{Zhou2018}.
\begin{equation}
    z = d,\;
    x = \frac{z (u - c_{x}) }{f_{x}},\;
    y = \frac{z (v - c_{y}) }{f_{y}}
\end{equation}

\noindent
where $u$ and $v$ are pixel coordinates, $c_x$ and $c_y$ are camera centers, $f_x$ and $f_y$ are focal lengths. Since both the SMPL models and the point clouds are in the camera coordinate system, they form paired data for training.

\subsubsection{Foreground Segmentation}
For each virtual character (subject), we compute the minimum 3D bounding boxes that enclose all its 3D keypoints. For simplicity, the sides of the bounding boxes are oriented parallel to world axes. As the 3D keypoints reside within the human body, we scale the bounding boxes by a factor of 1.2 in order to enclose the entire subject. Points inside the 3D bounding boxes are marked as foreground points. Note that cropping with bounding boxes inevitably includes points from the scene and other subjects. However, this problem also exists in real-life applications that leverage a top-down paradigm, making the study on GTA-Human-Point realistic and relevant.

\subsubsection{More Visualization}
We show additional examples from the GTA-Human-Point dataset in \Fig\ref{fig:gta_human++}. With a diverse collection of subjects, actions, and backgrounds, GTA-Human-Point includes interesting scenes where characters interact with the physical environment (from walking on the ground to falling off a building) and each other (resulting in severely incomplete point clouds due to heavy occlusion).

\ifCLASSOPTIONcaptionsoff
  \newpage
\fi




{
\bibliographystyle{IEEEtran}
\bibliography{references}

\begin{thebibliography}{10}
\providecommand{\url}[1]{#1}
\csname url@samestyle\endcsname
\providecommand{\newblock}{\relax}
\providecommand{\bibinfo}[2]{#2}
\providecommand{\BIBentrySTDinterwordspacing}{\spaceskip=0pt\relax}
\providecommand{\BIBentryALTinterwordstretchfactor}{4}
\providecommand{\BIBentryALTinterwordspacing}{\spaceskip=\fontdimen2\font plus
\BIBentryALTinterwordstretchfactor\fontdimen3\font minus
  \fontdimen4\font\relax}
\providecommand{\BIBforeignlanguage}[2]{{%
\expandafter\ifx\csname l@#1\endcsname\relax
\typeout{** WARNING: IEEEtran.bst: No hyphenation pattern has been}%
\typeout{** loaded for the language `#1'. Using the pattern for}%
\typeout{** the default language instead.}%
\else
\language=\csname l@#1\endcsname
\fi
#2}}
\providecommand{\BIBdecl}{\relax}
\BIBdecl

\bibitem{ofli2013berkeley}
F.~Ofli, R.~Chaudhry, G.~Kurillo, R.~Vidal, and R.~Bajcsy, ``Berkeley mhad: A
  comprehensive multimodal human action database,'' in \emph{2013 IEEE workshop
  on applications of computer vision (WACV)}.\hskip 1em plus 0.5em minus
  0.4em\relax IEEE, 2013, pp. 53--60.

\bibitem{Varol2017LearningFS}
G.~Varol, J.~Romero, X.~Martin, N.~Mahmood, M.~J. Black, I.~Laptev, and
  C.~Schmid, ``Learning from synthetic humans,'' \emph{2017 IEEE Conference on
  Computer Vision and Pattern Recognition (CVPR)}, pp. 4627--4635, 2017.

\bibitem{liu2021votehmr}
G.~Liu, Y.~Rong, and L.~Sheng, ``Votehmr: Occlusion-aware voting network for
  robust 3d human mesh recovery from partial point clouds,'' in
  \emph{Proceedings of the 29th ACM International Conference on Multimedia},
  2021, pp. 955--964.

\bibitem{loper2015smpl}
M.~Loper, N.~Mahmood, J.~Romero, G.~Pons-Moll, and M.~J. Black, ``{SMPL}: A
  skinned multi-person linear model,'' \emph{ACM transactions on graphics
  (TOG)}, vol.~34, no.~6, pp. 1--16, 2015.

\bibitem{kanazawa2018end}
A.~Kanazawa, M.~J. Black, D.~W. Jacobs, and J.~Malik, ``End-to-end recovery of
  human shape and pose,'' in \emph{Proceedings of the IEEE Conference on
  Computer Vision and Pattern Recognition}, 2018, pp. 7122--7131.

\bibitem{kolotouros2019learning}
N.~Kolotouros, G.~Pavlakos, M.~J. Black, and K.~Daniilidis, ``Learning to
  reconstruct 3d human pose and shape via model-fitting in the loop,'' in
  \emph{Proceedings of the IEEE/CVF International Conference on Computer
  Vision}, 2019, pp. 2252--2261.

\bibitem{kocabas2021pare}
M.~Kocabas, C.-H.~P. Huang, O.~Hilliges, and M.~J. Black, ``Pare: Part
  attention regressor for 3d human body estimation,'' \emph{arXiv preprint
  arXiv:2104.08527}, 2021.

\bibitem{li2022cliff}
Z.~Li, J.~Liu, Z.~Zhang, S.~Xu, and Y.~Yan, ``Cliff: Carrying location
  information in full frames into human pose and shape estimation,'' in
  \emph{European Conference on Computer Vision}.\hskip 1em plus 0.5em minus
  0.4em\relax Springer, 2022, pp. 590--606.

\bibitem{cai2022humman}
Z.~Cai, D.~Ren, A.~Zeng, Z.~Lin, T.~Yu, W.~Wang, X.~Fan, Y.~Gao, Y.~Yu, L.~Pan,
  F.~Hong, M.~Zhang, C.~C. Loy, L.~Yang, and Z.~Liu, ``Humman: Multi-modal 4d
  human dataset for versatile sensing and modeling,'' October 2022.

\bibitem{li2022lidarcap}
J.~Li, J.~Zhang, Z.~Wang, S.~Shen, C.~Wen, Y.~Ma, L.~Xu, J.~Yu, and C.~Wang,
  ``Lidarcap: Long-range marker-less 3d human motion capture with lidar point
  clouds,'' in \emph{Proceedings of the IEEE/CVF Conference on Computer Vision
  and Pattern Recognition}, 2022, pp. 20\,502--20\,512.

\bibitem{dai2022hsc4d}
Y.~Dai, Y.~Lin, C.~Wen, S.~Shen, L.~Xu, J.~Yu, Y.~Ma, and C.~Wang, ``Hsc4d:
  Human-centered 4d scene capture in large-scale indoor-outdoor space using
  wearable imus and lidar,'' in \emph{Proceedings of the IEEE/CVF Conference on
  Computer Vision and Pattern Recognition}, 2022, pp. 6792--6802.

\bibitem{bhatnagar2020combining}
B.~L. Bhatnagar, C.~Sminchisescu, C.~Theobalt, and G.~Pons-Moll, ``Combining
  implicit function learning and parametric models for 3d human
  reconstruction,'' in \emph{European Conference on Computer Vision}.\hskip 1em
  plus 0.5em minus 0.4em\relax Springer, 2020, pp. 311--329.

\bibitem{wang2021locally}
S.~Wang, A.~Geiger, and S.~Tang, ``Locally aware piecewise transformation
  fields for 3d human mesh registration,'' in \emph{Proceedings of the IEEE/CVF
  Conference on Computer Vision and Pattern Recognition}, 2021, pp. 7639--7648.

\bibitem{jiang2019skeleton}
H.~Jiang, J.~Cai, and J.~Zheng, ``Skeleton-aware 3d human shape reconstruction
  from point clouds,'' in \emph{Proceedings of the IEEE/CVF International
  Conference on Computer Vision}, 2019, pp. 5431--5441.

\bibitem{zhao2022lidar}
C.~Zhao, Y.~Ren, Y.~He, P.~Cong, H.~Liang, J.~Yu, L.~Xu, and Y.~Ma, ``Lidar-aid
  inertial poser: Large-scale human motion capture by sparse inertial and lidar
  sensors,'' \emph{arXiv preprint arXiv:2205.15410}, 2022.

\bibitem{shi2019skeleton}
L.~Shi, Y.~Zhang, J.~Cheng, and H.~Lu, ``Skeleton-based action recognition with
  multi-stream adaptive graph convolutional networks,'' \emph{arXiv preprint
  arXiv:1912.06971}, 2019.

\bibitem{newell2016stacked}
A.~Newell, K.~Yang, and J.~Deng, ``Stacked hourglass networks for human pose
  estimation,'' in \emph{European conference on computer vision}.\hskip 1em
  plus 0.5em minus 0.4em\relax Springer, 2016, pp. 483--499.

\bibitem{chen2018cascaded}
Y.~Chen, Z.~Wang, Y.~Peng, Z.~Zhang, G.~g. Yu, and J.~Sun, ``Cascaded pyramid
  network for multi-person pose estimation,'' in \emph{Proceedings of the IEEE
  Conference on Computer Vision and Pattern Recognition}, 2018, pp. 7103--7112.

\bibitem{chen2019hybrid}
K.~Chen, J.~Pang, J.~Wang, Y.~Xiong, X.~Li, S.~Sun, W.~Feng, Z.~Liu, J.~Shi,
  W.~Ouyang \emph{et~al.}, ``Hybrid task cascade for instance segmentation,''
  in \emph{Proceedings of the IEEE/CVF conference on computer vision and
  pattern recognition}, 2019, pp. 4974--4983.

\bibitem{sun2019hrnet}
K.~Sun, B.~Xiao, D.~Liu, and J.~Wang, ``Deep high-resolution representation
  learning for human pose estimation,'' in \emph{Proceedings of the IEEE/CVF
  Conference on Computer Vision and Pattern Recognition}, 2019, pp. 5693--5703.

\bibitem{qi2017pointnet++}
C.~R. Qi, L.~Yi, H.~Su, and L.~J. Guibas, ``Pointnet++: Deep hierarchical
  feature learning on point sets in a metric space,'' \emph{Advances in neural
  information processing systems}, vol.~30, 2017.

\bibitem{cai2021playing}
Z.~Cai, M.~Zhang, J.~Ren, C.~Wei, D.~Ren, J.~Li, Z.~Lin, H.~Zhao, S.~Yi,
  L.~Yang \emph{et~al.}, ``Playing for 3d human recovery,'' \emph{arXiv
  preprint arXiv:2110.07588}, 2021.

\bibitem{pavlakos2019expressive}
G.~Pavlakos, V.~Choutas, N.~Ghorbani, T.~Bolkart, A.~A. Osman, D.~Tzionas, and
  M.~J. Black, ``Expressive body capture: 3d hands, face, and body from a
  single image,'' in \emph{Proceedings of the IEEE/CVF Conference on Computer
  Vision and Pattern Recognition}, 2019, pp. 10\,975--10\,985.

\bibitem{osman2020star}
A.~A.~A. Osman, T.~Bolkart, and M.~J. Black, ``{STAR:} sparse trained
  articulated human body regressor,'' in \emph{{ECCV} {(6)}}, ser. Lecture
  Notes in Computer Science, vol. 12351.\hskip 1em plus 0.5em minus 0.4em\relax
  Springer, 2020, pp. 598--613.

\bibitem{xu2020ghum}
H.~Xu, E.~G. Bazavan, A.~Zanfir, W.~T. Freeman, R.~Sukthankar, and
  C.~Sminchisescu, ``Ghum \& ghuml: Generative 3d human shape and articulated
  pose models,'' in \emph{Proceedings of the IEEE/CVF Conference on Computer
  Vision and Pattern Recognition}, 2020, pp. 6184--6193.

\bibitem{bogo2016keep}
F.~Bogo, A.~Kanazawa, C.~Lassner, P.~Gehler, J.~Romero, and M.~J. Black, ``Keep
  it smpl: Automatic estimation of 3d human pose and shape from a single
  image,'' in \emph{European conference on computer vision}.\hskip 1em plus
  0.5em minus 0.4em\relax Springer, 2016, pp. 561--578.

\bibitem{pavlakos2018learning}
G.~Pavlakos, L.~Zhu, X.~Zhou, and K.~Daniilidis, ``Learning to estimate 3d
  human pose and shape from a single color image,'' in \emph{Proceedings of the
  IEEE Conference on Computer Vision and Pattern Recognition}, 2018, pp.
  459--468.

\bibitem{omran2018neural}
M.~Omran, C.~Lassner, G.~Pons-Moll, P.~Gehler, and B.~Schiele, ``Neural body
  fitting: Unifying deep learning and model based human pose and shape
  estimation,'' in \emph{2018 international conference on 3D vision
  (3DV)}.\hskip 1em plus 0.5em minus 0.4em\relax IEEE, 2018, pp. 484--494.

\bibitem{guler2019holopose}
R.~A. Guler and I.~Kokkinos, ``Holopose: Holistic 3d human reconstruction
  in-the-wild,'' in \emph{Proceedings of the IEEE/CVF Conference on Computer
  Vision and Pattern Recognition}, 2019, pp. 10\,884--10\,894.

\bibitem{kolotouros2019convolutional}
N.~Kolotouros, G.~Pavlakos, and K.~Daniilidis, ``Convolutional mesh regression
  for single-image human shape reconstruction,'' in \emph{Proceedings of the
  IEEE/CVF Conference on Computer Vision and Pattern Recognition}, 2019, pp.
  4501--4510.

\bibitem{li2020hybrik}
J.~Li, C.~Xu, Z.~Chen, S.~Bian, L.~Yang, and C.~Lu, ``Hybrik: {A} hybrid
  analytical-neural inverse kinematics solution for 3d human pose and shape
  estimation,'' in \emph{{CVPR}}.\hskip 1em plus 0.5em minus 0.4em\relax
  Computer Vision Foundation / {IEEE}, 2021, pp. 3383--3393.

\bibitem{georgakis2020hierarchical}
G.~Georgakis, R.~Li, S.~Karanam, T.~Chen, J.~Ko{\v{s}}eck{\'a}, and Z.~Wu,
  ``Hierarchical kinematic human mesh recovery,'' in \emph{European Conference
  on Computer Vision}.\hskip 1em plus 0.5em minus 0.4em\relax Springer, 2020,
  pp. 768--784.

\bibitem{luo20203d}
Z.~Luo, S.~A. Golestaneh, and K.~M. Kitani, ``3d human motion estimation via
  motion compression and refinement,'' in \emph{Proceedings of the Asian
  Conference on Computer Vision}, 2020.

\bibitem{sun2020monocular}
Y.~Sun, Q.~Bao, W.~Liu, Y.~Fu, B.~Michael~J., and T.~Mei, ``Monocular,
  one-stage, regression of multiple 3d people,'' in \emph{ICCV}, October 2021.

\bibitem{dwivedi2021learning}
S.~K. Dwivedi, N.~Athanasiou, M.~Kocabas, and M.~J. Black, ``Learning to
  regress bodies from images using differentiable semantic rendering,'' in
  \emph{Proceedings of the IEEE/CVF International Conference on Computer
  Vision}, 2021, pp. 11\,250--11\,259.

\bibitem{kolotouros2021probabilistic}
N.~Kolotouros, G.~Pavlakos, D.~Jayaraman, and K.~Daniilidis, ``Probabilistic
  modeling for human mesh recovery,'' in \emph{Proceedings of the IEEE/CVF
  International Conference on Computer Vision}, 2021, pp. 11\,605--11\,614.

\bibitem{rajasegaran2022tracking}
J.~Rajasegaran, G.~Pavlakos, A.~Kanazawa, and J.~Malik, ``Tracking people by
  predicting 3d appearance, location and pose,'' in \emph{Proceedings of the
  IEEE/CVF Conference on Computer Vision and Pattern Recognition}, 2022, pp.
  2740--2749.

\bibitem{sun2022putting}
Y.~Sun, W.~Liu, Q.~Bao, Y.~Fu, T.~Mei, and M.~J. Black, ``Putting people in
  their place: Monocular regression of 3d people in depth,'' in
  \emph{Proceedings of the IEEE/CVF Conference on Computer Vision and Pattern
  Recognition}, 2022, pp. 13\,243--13\,252.

\bibitem{wang2023zolly}
W.~Wang, Y.~Ge, H.~Mei, Z.~Cai, Q.~Sun, Y.~Wang, C.~Shen, L.~Yang, and
  T.~Komura, ``Zolly: Zoom focal length correctly for perspective-distorted
  human mesh reconstruction,'' \emph{arXiv preprint arXiv:2303.13796}, 2023.

\bibitem{kanazawa2019learning}
A.~Kanazawa, J.~Y. Zhang, P.~Felsen, and J.~Malik, ``Learning 3d human dynamics
  from video,'' in \emph{Proceedings of the IEEE/CVF Conference on Computer
  Vision and Pattern Recognition}, 2019, pp. 5614--5623.

\bibitem{sun2019human}
Y.~Sun, Y.~Ye, W.~Liu, W.~Gao, Y.~Fu, and T.~Mei, ``Human mesh recovery from
  monocular images via a skeleton-disentangled representation,'' in
  \emph{Proceedings of the IEEE/CVF International Conference on Computer
  Vision}, 2019, pp. 5349--5358.

\bibitem{mehta2020xnect}
D.~Mehta, O.~Sotnychenko, F.~Mueller, W.~Xu, M.~Elgharib, P.~Fua, H.-P. Seidel,
  H.~Rhodin, G.~Pons-Moll, and C.~Theobalt, ``Xnect: Real-time multi-person 3d
  motion capture with a single rgb camera,'' \emph{ACM Transactions on Graphics
  (TOG)}, vol.~39, no.~4, pp. 82--1, 2020.

\bibitem{moon2020i2l}
G.~Moon and K.~M. Lee, ``I2l-meshnet: Image-to-lixel prediction network for
  accurate 3d human pose and mesh estimation from a single {RGB} image,'' in
  \emph{{ECCV} {(7)}}, ser. Lecture Notes in Computer Science, vol.
  12352.\hskip 1em plus 0.5em minus 0.4em\relax Springer, 2020, pp. 752--768.

\bibitem{kocabas2020vibe}
M.~Kocabas, N.~Athanasiou, and M.~J. Black, ``Vibe: Video inference for human
  body pose and shape estimation,'' in \emph{Proceedings of the IEEE/CVF
  Conference on Computer Vision and Pattern Recognition}, 2020, pp. 5253--5263.

\bibitem{choi2021beyond}
H.~Choi, G.~Moon, and K.~M. Lee, ``Beyond static features for temporally
  consistent 3d human pose and shape from a video,'' in \emph{Conference on
  Computer Vision and Pattern Recognition (CVPR)}, 2021.

\bibitem{shotton2011real}
J.~Shotton, A.~Fitzgibbon, M.~Cook, T.~Sharp, M.~Finocchio, R.~Moore,
  A.~Kipman, and A.~Blake, ``Real-time human pose recognition in parts from
  single depth images,'' in \emph{CVPR 2011}.\hskip 1em plus 0.5em minus
  0.4em\relax Ieee, 2011, pp. 1297--1304.

\bibitem{haque2016towards}
A.~Haque, B.~Peng, Z.~Luo, A.~Alahi, S.~Yeung, and L.~Fei-Fei, ``Towards
  viewpoint invariant 3d human pose estimation,'' in \emph{European conference
  on computer vision}.\hskip 1em plus 0.5em minus 0.4em\relax Springer, 2016,
  pp. 160--177.

\bibitem{xiong2019a2j}
F.~Xiong, B.~Zhang, Y.~Xiao, Z.~Cao, T.~Yu, J.~T. Zhou, and J.~Yuan, ``A2j:
  Anchor-to-joint regression network for 3d articulated pose estimation from a
  single depth image,'' in \emph{Proceedings of the IEEE/CVF International
  Conference on Computer Vision}, 2019, pp. 793--802.

\bibitem{zhou2020learning}
Y.~Zhou, H.~Dong, and A.~El~Saddik, ``Learning to estimate 3d human pose from
  point cloud,'' \emph{IEEE Sensors Journal}, vol.~20, no.~20, pp.
  12\,334--12\,342, 2020.

\bibitem{garau2021deca}
N.~Garau, N.~Bisagno, P.~Br{\'o}dka, and N.~Conci, ``Deca: Deep
  viewpoint-equivariant human pose estimation using capsule autoencoders,'' in
  \emph{Proceedings of the IEEE/CVF International Conference on Computer
  Vision}, 2021, pp. 11\,677--11\,686.

\bibitem{bashirov2021real}
R.~Bashirov, A.~Ianina, K.~Iskakov, Y.~Kononenko, V.~Strizhkova, V.~Lempitsky,
  and A.~Vakhitov, ``Real-time rgbd-based extended body pose estimation,'' in
  \emph{Proceedings of the IEEE/CVF Winter Conference on Applications of
  Computer Vision}, 2021, pp. 2807--2816.

\bibitem{wang2020sequential}
K.~Wang, J.~Xie, G.~Zhang, L.~Liu, and J.~Yang, ``Sequential 3d human pose and
  shape estimation from point clouds,'' in \emph{Proceedings of the IEEE/CVF
  Conference on Computer Vision and Pattern Recognition}, 2020, pp. 7275--7284.

\bibitem{qi2017pointnet}
C.~R. Qi, H.~Su, K.~Mo, and L.~J. Guibas, ``Pointnet: Deep learning on point
  sets for 3d classification and segmentation,'' in \emph{Proceedings of the
  IEEE conference on computer vision and pattern recognition}, 2017, pp.
  652--660.

\bibitem{wang2019dynamic}
Y.~Wang, Y.~Sun, Z.~Liu, S.~E. Sarma, M.~M. Bronstein, and J.~M. Solomon,
  ``Dynamic graph cnn for learning on point clouds,'' \emph{Acm Transactions On
  Graphics (tog)}, vol.~38, no.~5, pp. 1--12, 2019.

\bibitem{qin2020bipointnet}
H.~Qin, Z.~Cai, M.~Zhang, Y.~Ding, H.~Zhao, S.~Yi, X.~Liu, and H.~Su,
  ``Bipointnet: Binary neural network for point clouds,'' \emph{arXiv preprint
  arXiv:2010.05501}, 2020.

\bibitem{li2019selective}
X.~Li, W.~Wang, X.~Hu, and J.~Yang, ``Selective kernel networks,'' in
  \emph{Proceedings of the IEEE/CVF conference on computer vision and pattern
  recognition}, 2019, pp. 510--519.

\bibitem{shen2021efficient}
Z.~Shen, M.~Zhang, H.~Zhao, S.~Yi, and H.~Li, ``Efficient attention: Attention
  with linear complexities,'' in \emph{Proceedings of the IEEE/CVF winter
  conference on applications of computer vision}, 2021, pp. 3531--3539.

\bibitem{vaswani2017attention}
A.~Vaswani, N.~Shazeer, N.~Parmar, J.~Uszkoreit, L.~Jones, A.~N. Gomez,
  {\L}.~Kaiser, and I.~Polosukhin, ``Attention is all you need,''
  \emph{Advances in neural information processing systems}, vol.~30, 2017.

\bibitem{groueix20183d}
T.~Groueix, M.~Fisher, V.~G. Kim, B.~C. Russell, and M.~Aubry, ``3d-coded: 3d
  correspondences by deep deformation,'' in \emph{Proceedings of the European
  Conference on Computer Vision (ECCV)}, 2018, pp. 230--246.

\bibitem{fabbri2018learning}
M.~Fabbri, F.~Lanzi, S.~Calderara, A.~Palazzi, R.~Vezzani, and R.~Cucchiara,
  ``Learning to detect and track visible and occluded body joints in a virtual
  world,'' in \emph{Proceedings of the European Conference on Computer Vision
  (ECCV)}, 2018, pp. 430--446.

\bibitem{cao2020long}
Z.~Cao, H.~Gao, K.~Mangalam, Q.-Z. Cai, M.~Vo, and J.~Malik, ``Long-term human
  motion prediction with scene context,'' in \emph{European Conference on
  Computer Vision}.\hskip 1em plus 0.5em minus 0.4em\relax Springer, 2020, pp.
  387--404.

\bibitem{Zhou2018}
Q.-Y. Zhou, J.~Park, and V.~Koltun, ``{Open3D}: {A} modern library for {3D}
  data processing,'' \emph{arXiv:1801.09847}, 2018.

\end{thebibliography}
}
\begin{IEEEbiography}[{\includegraphics[width=1in,height=1.25in,clip,keepaspectratio]{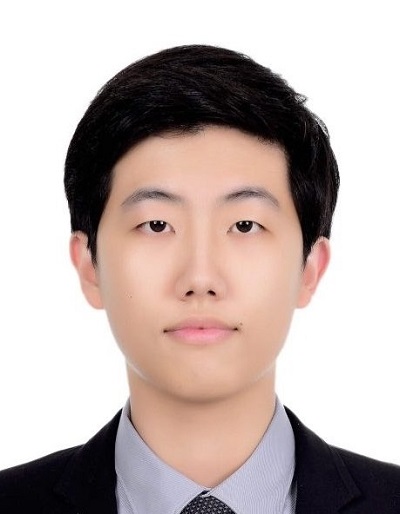}}]{Zhongang Cai}
  is currently a Year 3 Ph.D. Student at Nanyang Technological University (NTU) supervised by Prof Liu Ziwei and Prof Chen Change Loy, and a Senior Algorithm Researcher at SenseTime International Pte. Ltd. He attained his bachelor’s degree at Nanyang Technological University and was awarded the Lee Kuan Yew Gold Medal as a top student. His research interest includes point clouds and virtual humans. To date, he has published 10+ papers on top venues such as ICLR, CVPR, ICCV and ECCV. He also serves as a reviewer for top machine learning and computer vision conferences, such as NeurIPS, ICLR and ICCV.
\end{IEEEbiography}

\begin{IEEEbiography}[{\includegraphics[width=1in,height=1.25in,clip,keepaspectratio]{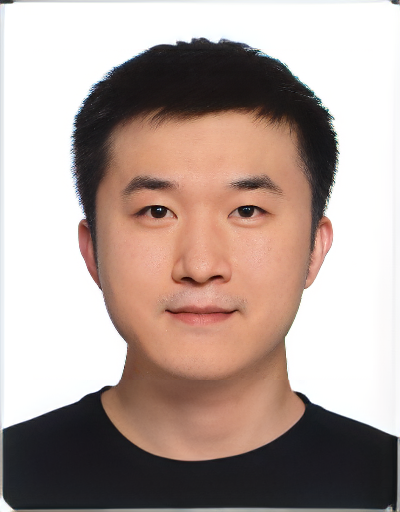}}]{Liang Pan}
    received the PhD degree in Mechanical Engineering from National University of Singapore (NUS) in 2019.
    He is currently a Research Fellow at S-Lab, Nanyang Technological University (NTU).  
    Previously, He is a Research Fellow at the Advanced Robotics Centre from National University of Singapore.
    His research interests include computer vision and
    3D point cloud, with focus on shape analysis, deep learning, and 3D human. 
    He also serves as a reviewer for top computer vision and robotics conferences, such as CVPR, ICCV, ECCV, ICML, NeurlPS, and ICLR.
\end{IEEEbiography}

\begin{IEEEbiography}[{\includegraphics[width=1in,height=1.25in,clip,keepaspectratio]{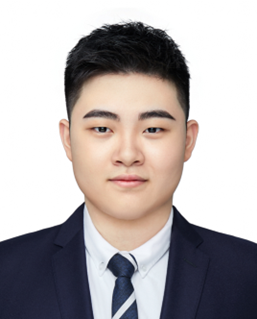}}]{Chen Wei}
  graduated from Nanyang Technological University. He was an algorithm research intern at S-Lab @ NTU in 2021 and continued part-time internship at SenseTime International Pte. Ltd in 2022. His work focuses on the generation of synthetic human data from game engines.
\end{IEEEbiography}

\begin{IEEEbiography}[{\includegraphics[width=1in,height=1.25in,clip,keepaspectratio]{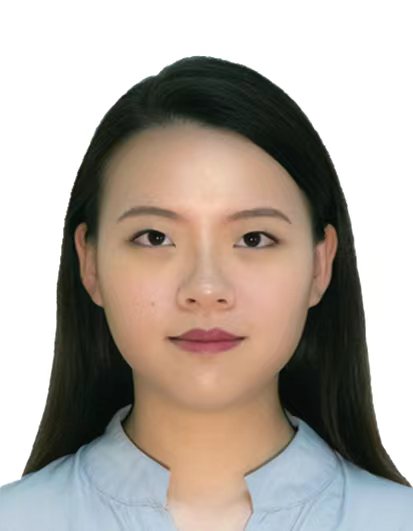}}]{Wanqi Yin} received the Master's degree in Precision Engineering from the University of Tokyo in 2020. She is currently an Algorithm Researcher at SenseTime International Pte. Ltd. Her research interests include computer vision and 3D measurement, with a focus on motion capture, virtual humans, and multi-camera system.
\end{IEEEbiography}

\begin{IEEEbiography}[{\includegraphics[width=1in,height=1.25in,clip,keepaspectratio]{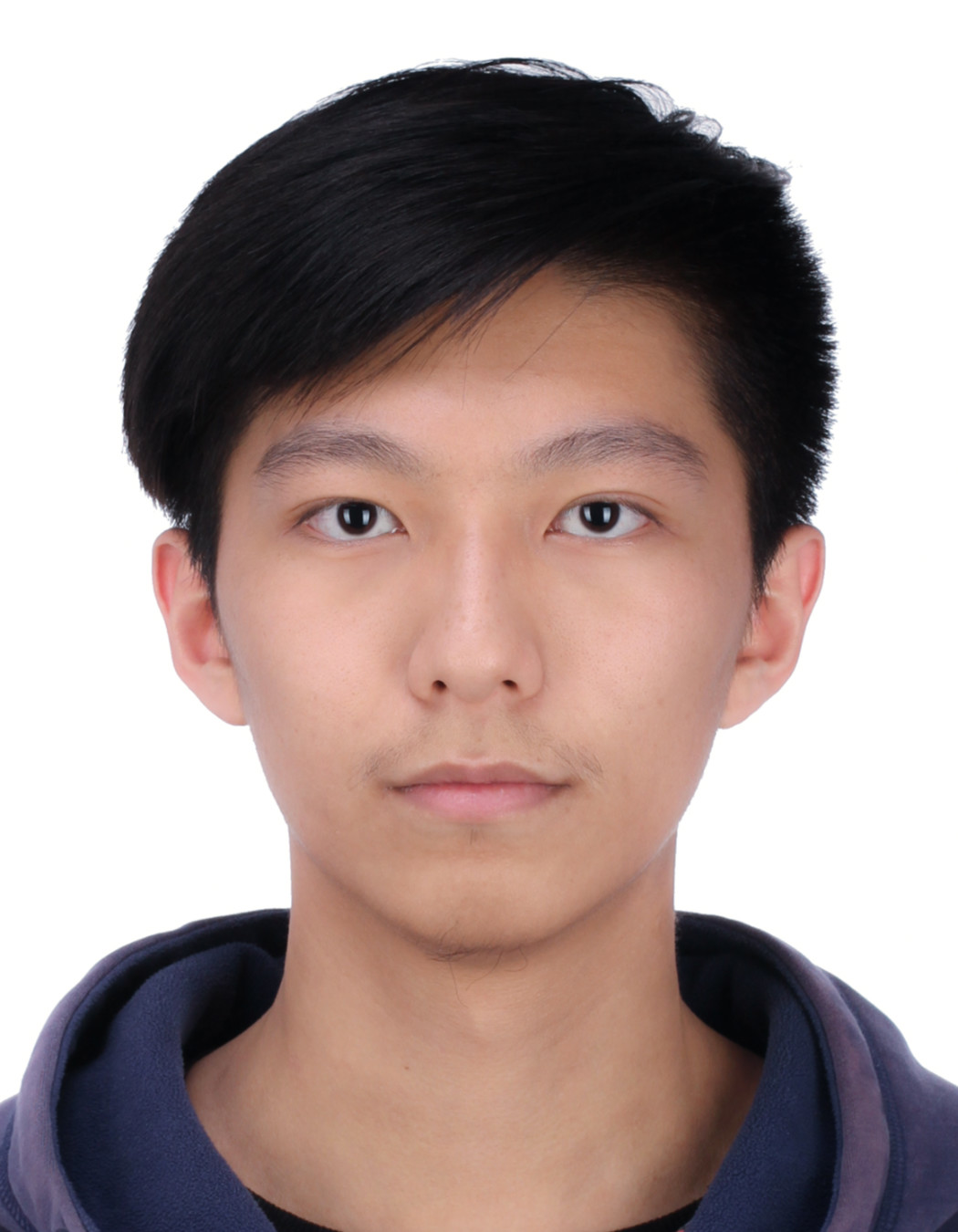}}]{Fangzhou Hong}  received the BEng degree in Software Engineering from Tsinghua University, China, in 2020. He is currently a Ph.D. student in the School of Computer Science and Engineering at Nanyang Technological University. His research interests lie on the computer vision and deep learning. Particularly he is interested in 3D representation learning.
\end{IEEEbiography}

\begin{IEEEbiography}[{\includegraphics[width=1in,height=1.25in,clip,keepaspectratio]{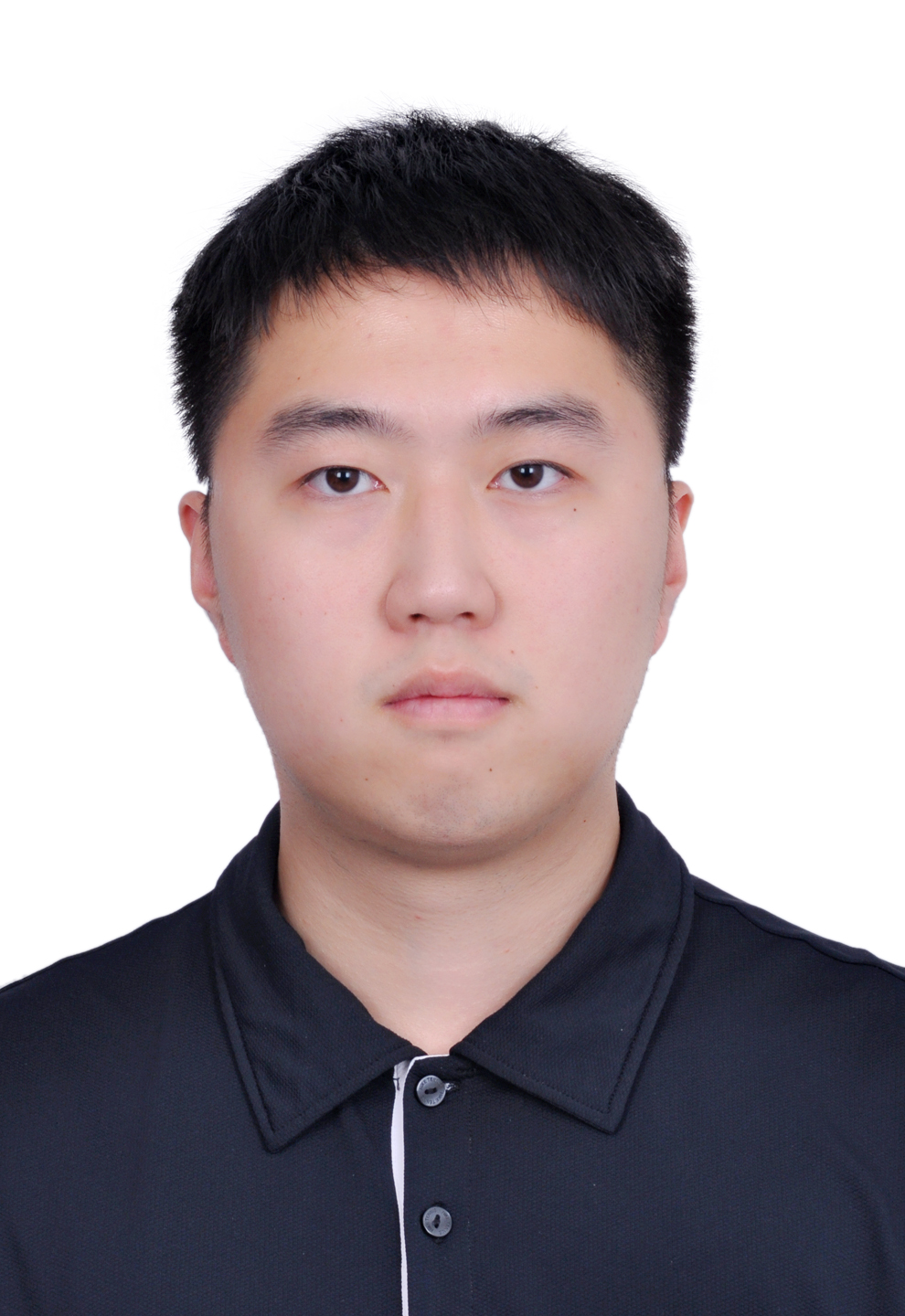}}]{Mingyuan Zhang}
  received a B.S. degree in computer science and engineering from Beihang University, China. He is currently pursuing a Ph.D. degree at MMLab@NTU, advised by Prof. Ziwei Liu. His research interests in computer vision include motion synthesis, 3D pose estimation, and scene understanding. He has published papers on top-tier conferences, including CVPR, ECCV, ICCV, ICLR, and AAAI.
\end{IEEEbiography}

\begin{IEEEbiography}[{\includegraphics[width=1in,height=1.25in,clip,keepaspectratio]{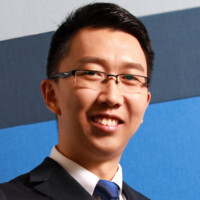}}]{Chen Change Loy}
 (Senior Member, IEEE) received the PhD degree in computer science from the Queen Mary University of London, in 2010. He is an associate professor with the School of Computer Science and Engineering, Nanyang Technological University. Prior to joining NTU, he served as a research assistant professor with the Department of Information Engineering, The Chinese University of Hong Kong, from 2013 to 2018. His research interests include computer vision and deep learning with a focus on image/video restoration and enhancement, generative tasks, and representation learning. He serves as an Associate Editor of the International Journal of Computer Vision (IJCV) and IEEE Transactions on Pattern Analysis and Machine Intelligence (TPAMI). He also serves/served as an Area Chair of ICCV 2021, CVPR (2021, 2019), ECCV (2022, 2018), AAAI (2021-2023), and BMVC (2018-2021).
\end{IEEEbiography}

\begin{IEEEbiography}[{\includegraphics[width=1in,height=1.25in,clip,keepaspectratio]{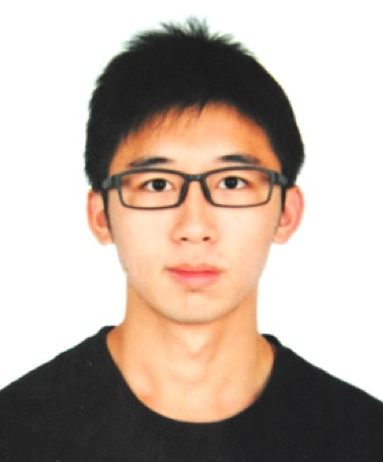}}]{Lei Yang}
  is currently a Research Director at SenseTime Group Inc., under the supervision of Prof. Xiaogang Wang. Prior to that, Lei  received his Ph.D. degree from the Chinese University of Hong Kong in 2020, advised by Prof. Dahua Lin and Prof. Xiaoou Tang. Before that, Lei obtained his B.E. degree from Tsinghua University in 2015. He has published over 10 papers on top conferences in relevant fields, including CVPR, ICCV, ECCV, AAAI, SIGGRAPH and RSS.
\end{IEEEbiography}

\begin{IEEEbiography}[{\includegraphics[width=1in,height=1.25in,clip,keepaspectratio]{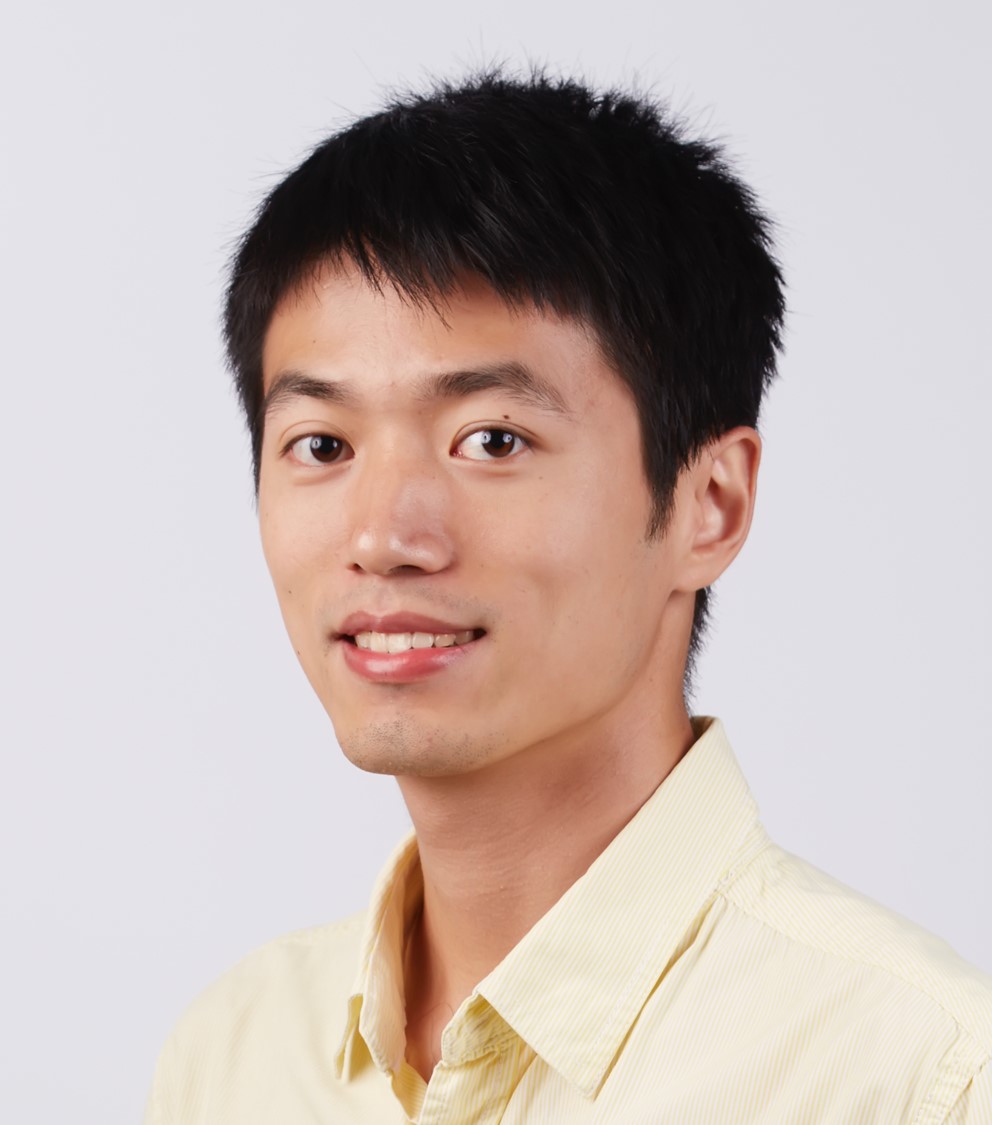}}]{Ziwei Liu}
  is currently an Assistant Professor at Nanyang Technological University, Singapore. Previously, he was a senior research fellow at the Chinese University of Hong Kong and a postdoctoral researcher at University of California, Berkeley. Ziwei received his PhD from the Chinese University of Hong Kong. His research revolves around computer vision, machine learning and computer graphics. He has published extensively on top-tier conferences and journals in relevant fields, including CVPR, ICCV, ECCV, NeurIPS, ICLR, ICML, TPAMI, TOG and Nature - Machine Intelligence. He is the recipient of Microsoft Young Fellowship, Hong Kong PhD Fellowship, ICCV Young Researcher Award and HKSTP Best Paper Award. He also serves as an Area Chair of ICCV, NeurIPS and ICLR.
\end{IEEEbiography}

\vfill

\end{document}